\documentclass[]{elsarticle}
\usepackage[utf8]{inputenc}
\usepackage{natbib}
\usepackage{csquotes}
\usepackage{times}
\usepackage{latexsym}
\usepackage{comment}
\usepackage{hyperref}
\usepackage{geometry}
\usepackage{graphicx}
\usepackage{multirow}
\usepackage{hhline}
\usepackage{amsmath}
\usepackage[toc,page]{appendix}
\usepackage{float}

\usepackage{xcolor}
\usepackage[font=small]{caption}

\author{
Christel Chappuis$^1$,
Eliot Walt$^2$, 
Vincent Mendez$^3$,
Sylvain Lobry$^4$,
Bertrand Le Saux$^5$,
Devis Tuia$^1$
}
\address{$^1$ Environmental Computer Science and Earth Observation Laboratory, École Polytechnique Fédérale de Lausanne, Sion, Switzerland\\
$^2$ École Polytechnique Fédérale de Lausanne, Switzerland\\
$^3$ Neuro X Institute, École Polytechnique Fédérale de Lausanne, Genève, Switzerland\\
$^4$ LIPADE, Université Paris Cit\'e, Paris, France\\
$^5$ European Space Agency $\Phi$lab, Frascati, Italy
}

\date{\today}

\title{The curse of language biases in remote sensing VQA:\\
the role of spatial attributes, language diversity, and the need for clear evaluation}

\begin{document}

\begin{abstract}
Remote sensing visual question answering (RSVQA) opens new opportunities for the use of overhead imagery by the general public, by enabling human-machine interaction with natural language. Building on the recent advances in natural language processing and computer vision, the goal of RSVQA is to answer a question formulated in natural language about a remote sensing image. 
Language understanding is essential to the success of the task, but has not yet been thoroughly examined in RSVQA. 
In particular, the problem of language biases is often overlooked in the remote sensing community, which can impact model robustness and lead to wrong conclusions about the performances of the model.
Thus, the present work aims at highlighting the problem of language biases in RSVQA with a threefold analysis strategy: visual blind models, adversarial testing and dataset analysis. This analysis focuses both on model and data. Moreover, we motivate the use of more informative and complementary evaluation metrics sensitive to the issue. We compare a recurrent neural network (Skip-thoughts) used in early RSVQA works with an attention-based Transformer (distilBERT), as well as training strategies (frozen or fine-tuned language models). 
The gravity of language biases in RSVQA is then exposed for all of these methods with the training of models discarding the image data and the manipulation of the visual input during inference. Finally, a detailed analysis of question-answer distribution demonstrates the root of the problem in the data itself.
Thanks to this analytical study, we observed that biases in remote sensing are more severe than in standard VQA, likely due to the specifics of existing remote sensing datasets for the task, e.g. geographical similarities and sparsity, as well as a simpler vocabulary and question generation strategies. 
While new, improved and less-biased datasets appear as a necessity for the development of the promising field of RSVQA, we demonstrate that more informed, relative evaluation metrics remain much needed to transparently communicate results of future RSVQA methods. 
\end{abstract}

\maketitle

\section{Introduction}
Giving access to information contained in remote sensing images by means of queries entered in natural language has the potential of widening the use of remote sensing imagery. 
Large quantities of Earth Observation data from remote sensing satellite missions are stored, but only a fraction is effectively used~\citep{lachezar_challenges_2018}.
Facilitating the extraction of information useful for applications in environmental monitoring, agriculture, or urban planning could increase benefits derived from the abundance of data sensed~\citep{campsvalls_deep_2021}.
Such extraction would need to be achievable in a simple way, for instance as a query formulated in a humanly intelligible manner. Remote Sensing Visual Question Answering (RSVQA~\citep{lobry_rsvqa_2020}) aims to tackle this challenge, building on deep learning, computer vision and natural language processing (NLP), and brings new opportunities in applied environmental sciences~\citep{tuia_towards_2021}. Since its proposition, RSVQA has seen expanding research in terms of model architecture~\citep{felix_cross-modal_2021,yuan_easy_2022, bazi_bi-modal_2022, chappuis_prompt-rsvqa_2022}, bi-modal image/text fusion~\citep{chappuis_how_2021, siebert_multi-modal_2022} or datasets~\citep{lobry_rsvqa_2020, lobry_rsvqa_2021, zheng_mutual_2021, faure_embedding_2022}, but the role of NLP in the RSVQA balance, as well as biases in the datasets has so far received less attention than the visual part of models. However, understanding these biases has been crucial to counteract and develop more balanced VQA datasets and successful models in computer vision~\citep{goyal_making_2017, agrawal_dont_2018, niu_counterfactual_2021}, as well as avoid over-promising solution or discriminating, harmful algorithms. 

The goal of this study is to fill this gap and assess in details the language component of RSVQA, in particular the challenge of language biases, both in the method and in the datasets. Based on these findings, we encourage new relative and straightforward evaluation metrics considering this issue. The language bias analysis method is threefold (visual blind models, adversarial testing and dataset analysis) and is performed on a set of models in which a variety of language models and fusion strategies are compared, with or without fine-tuning. We focus on language models with the Recurrent Neural Network (RNN) architecture, illustrated by a Skip-thoughts RNN~\citep{kiros_skip-thought_2015}, and the attention-based language Transformers represented by the BERT-family (BERT~\citep{devlin_bert_2019}, distilBERT~\citep{sanh_distilbert_2020}, roBERTa~\citep{liu_roberta_2019}).
The first analysis method considers the reliance of RSVQA to the language component: to do so, we alter the RSVQA models by removing the visual input. These visually impaired models, referred to as ``visual blind'' in this study (as suggested in~\citet{chappuis_prompt-rsvqa_2022}), consider only the questions without any information from the images. They are trained with and without fine-tuning of the language model and compared to their corresponding full version, i.e. the RSVQA model with access to visual information from the images. This experiment highlights the considerable amount of information hidden in the unbalanced distribution of questions and answers, and the opportunity for language models to take short-cuts by learning the most probable answer given the words of the question.
Secondly, we use an adversarial testing procedure to evaluate if the reference models are sensitive to disruptions in the visual input. During inference, we modify the image and observe how the model reacts in terms of performance, compared to the initial results of the unconstrained RSVQA model and those of the visual blind approach.
Thirdly and seeking to better understand RSVQA datasets, a detailed data analysis is conducted to practically identify the aspects that lead to language biases. Our manual analysis procedure focuses on the distribution of the question-answer pairs in several datasets from the remote sensing literature, with a focus on the low resolution dataset of RSVQA, whose generation was inspired by the CLEVR protocol~\citep{johnson_clevr_2017}.

The main contribution of this analytical study is to provide a clear understanding on the biases in RSVQA, their importance and a strategy to transparently evaluate models considering this issue. In the RSVQA task, samples are not composed of 2 (i.e. input and target) but 3 elements (i.e. image input, question input and target answer), leaving the possibility for models to use short-cuts in the question-answer distribution. As training, validation and testing splits of a dataset are often generated with the same automatic procedure in RSVQA datasets, we show that models tend to overfit by learning these short-cuts. 
While this problem was already identified in classical VQA~\citep{goyal_making_2017, agrawal_dont_2018, kervadec_roses_2021}, it remains to be recognized in RSVQA. In fact, we show that it is even more crucial in RSVQA, accentuated by the characteristics of remote sensing and datasets development strategies. 
The lower language diversity, with questions generated from simple templates and about a limited amount of objects and relations, certainly contributes to making the biases stronger. In addition to language aspects, images in remote sensing are also not as diverse and accessible as in natural vision. With a bird-eye perspective, the theme diversity is limited and the resolution can be unconventional and restrictive for the human eye. Furthermore, for RSVQA datasets covering a specific geographical area, bias can be a natural component, as the image content is typically dependent on the region considered. 
Performances of RSVQA models are subject to all these challenges. While the displayed accuracy might be high, there is no guarantee that the predicted answers are indeed grounded in the visual input. Thus, to continue experimenting on existing strongly biased datasets, an adequate, relative evaluation metric is key. We propose two such metrics in this paper, hoping that this type of evaluation pratices will become more common in future research.
In the longer term and to advance robust research in RSVQA, we believe new data designed with bias reduction in mind will be important in order to avoid the risk of stagnating into cycles of methodological improvement on biased datasets. This analytical study of existing datasets is thus crucial to improve awareness on the challenge of language biases in the remote sensing community.


The rest of the paper is organized in the following manner. Section~\ref{sec:RW} reviews relevant related works. Next, the analysis methods are presented in Section~\ref{sec:methods}. This section starts with a description of the baseline architecture and the different language models used in the analytical study. Our main contribution is an analysis strategy composed of three parts: the ``visual blind'' models, the adversarial testing and the remote sensing VQA dataset analysis. Then, Section~\ref{sec:evalMetrics} is dedicated to another contribution: the relative evaluation metrics. These metrics are proposed to transparently consider the problem of language biases in results reporting. Section~\ref{sec:DA} presents the remote sensing VQA datasets examined. The results are displayed and discussed together in Section~\ref{sec:results}. Finally, a higher-level discussion and the main take-away messages are summarized in Section~\ref{sec:discussion_conclusion}.

\section{Related work}
\label{sec:RW}

\paragraph{Visual Question Answering}
The task of RSVQA~\citep{lobry_rsvqa_2020} was inspired by the one of Visual Question Answering (VQA)~\citep{antol_vqa_2015} that aims at retrieving an answer to a question about a natural image. Since the initial VQA proposition, considerable developments have been witnessed in classical VQA. 
Some methods use attention mechanisms~\citep{yang_stacked_2016, anderson_bottom-up_2018}, while others decompose the search for the answer into a series of simple actions (neural module network~\citep{andreas_neural_2016}), integrate external knowledge~\citep{wu_image_2018} or use graph representations~\citep{teney_graph-structured_2017} to enhance a closer representation of both modalities.
More recently, Transformers architectures using large-scale pre-training, also referred to as foundation models, demonstrate state-of-the-art performances on multiple tasks, including VQA. Foundation models originate in the NLP field with models such as BERT~\citep{devlin_bert_2019} or GPT-3~\citep{brown_language_2020}, and have received much attention in the vision-and-language field. Besides VQA, foundation models have also been used in tasks such as captioning~\citep{yu_coca_2022}, retrieval~\citep{cao_image-text_2022} and commonsense knowledge~\citep{bhargava_commonsense_2022}.
For example, UNITER~\citep{chen_uniter_2020} was pre-trained on a large set of data from four datasets: MS-COCO~\citep{fleet_microsoft_2014}, Visual Genome~\citep{krishna_visual_2017}, Conceptual Captions~\citep{sharma_conceptual_2018} and SBU captions~\citep{Ordonez_Im2Text_2011} and using 4 different pre-training tasks: masking words in the texts, regions in the images, matching the two modalities or aligning words and regions.
An architecture able to consider uni- and multi-modal data (text or image, or text and image) was proposed by~\citep{wang_ufo_2021} and trained on the same set of datasets. 
Going beyond static images and adding the temporal dimension (e.g. videos) is the focus of MERLOT~\citep{zellers_merlot_2021}, which was trained on 6 million YouTube videos with transcriptions. Interestingly, while it showed strong out-of-the-box temporal representations and commonsense reasoning capacities, it also performed well on static images like in the VQA task. Finally, text and image data collected from the web (900 million samples dataset) was used to train Florence~\citep{yuan_florence_2021}. This foundation model is meant to be a vision foundation model, pre-trained and able to adapt and solve vision tasks across spatial and temporal dimension with zero- or few-shot capabilities.

\paragraph{Dataset biases} Despite all these successful examples, it has been shown that VQA models trained or fine-tuned on VQA datasets are able to leverage the distribution of question and answer pairs instead of retrieving the answer from the visual field as intended. This issue is usually referred to as language biases, and affects the robustness and trustworthiness of VQA models. Practically, it means that models tend to learn the most common answer relative to some elements of the question without grounding the search in the image.
In classical VQA, several contributions have highlighted this challenge with priors (most common answer for the entire training set or subsets) and language-only models (``visual blind''). Strategies have been proposed to overcome these biases~\citep{goyal_making_2017, agrawal_dont_2018, kervadec_roses_2021}. 
In fact, \citet{antol_vqa_2015} anticipated this issue by analysing their dataset, VQA, with human subjects and models being provided the question, the image or the image caption.
\citet{goyal_making_2017} released an augmented version of the VQA dataset~\citep{antol_vqa_2015} by associating a second, similar image for each question with a different answer. The resulting dataset, VQA v2.0, can be considered the balanced version of its predecessor. \citet{agrawal_dont_2018} suggested adaptions in the data as well by defining new splits with different priors distributions, and publish a new architecture restricted by concepts of visual grounding. Also tackling methodology, \citet{niu_counterfactual_2021} proposed a counterfactual inference framework aiming at removing the effect of language biases by subtracting language-only predictions from the predictions of the full architecture. 
Evaluation metrics represent yet another strategy to tackle language bias. \citet{kervadec_roses_2021} compared performances of rare and frequent question-answer pairs to conclude that using out-of-distribution performance (rare pairs) is a better metric to evaluate the reasoning abilities of a method. Finally, looking at the other modality, vision, VQA models also tend to use irrelevant information in the image to provide the answer~\citep{gupta_swapmix_2022}. Referred to as ``over-reliance on visual context'', this has also been highlighted as a vulnerability of classical VQA models.

\paragraph{Remote Sensing Visual Question Answering}
Following the initial RSVQA proposition~\citep{lobry_rsvqa_2020} composed of a dataset and a baseline architecture, several other datasets have been published: FloodNet~\citep{rahnemoonfar_floodnet_2020} is concerned with question answering relative to a flooding natural disaster, RSIVQA~\citep{zheng_mutual_2021} is composed of existing classification and object detection datasets, RSVQAxBEN~\citep{lobry_rsvqa_2021} is a large-scale dataset focusing on land cover questions, and CDVQA~\citep{yuan_change_2022} deals with change between bi-temporal images. Spatial relations are investigated in~\citet{faure_embedding_2022}. Methodologically, an architecture using bi-modal attention mechanism was presented in~\citet{zheng_mutual_2021}, while bi-modal fusion strategies in the baseline architecture were studied in~\citet{chappuis_how_2021}. 
In parallel, \citet{yuan_easy_2022} proposed a self-paced curriculum learning approach to gradually train the model from easier to harder questions. \citet{lobry_better_2020} tackled in details the challenging question type of ``Counting'' with a multi-head strategy combining classification and regression, whereas \citet{felix_cross-modal_2021} achieved good results especially on this question type by integrating an object detection visual feature extractor in their method. Visual transformers have been recently integrated in RSVQA architectures, either as image feature extractor~\citep{bazi_bi-modal_2022}, or as a fusion strategy~\citep{siebert_multi-modal_2022}. Finally, Prompt-RSVQA~\citep{chappuis_prompt-rsvqa_2022, christelmveo} suggests to describe the image with words and retrieve the answer with a single language transformer model fed with visual context.

\begin{figure}[!t]
    \centering
    \includegraphics[width =0.95\textwidth]{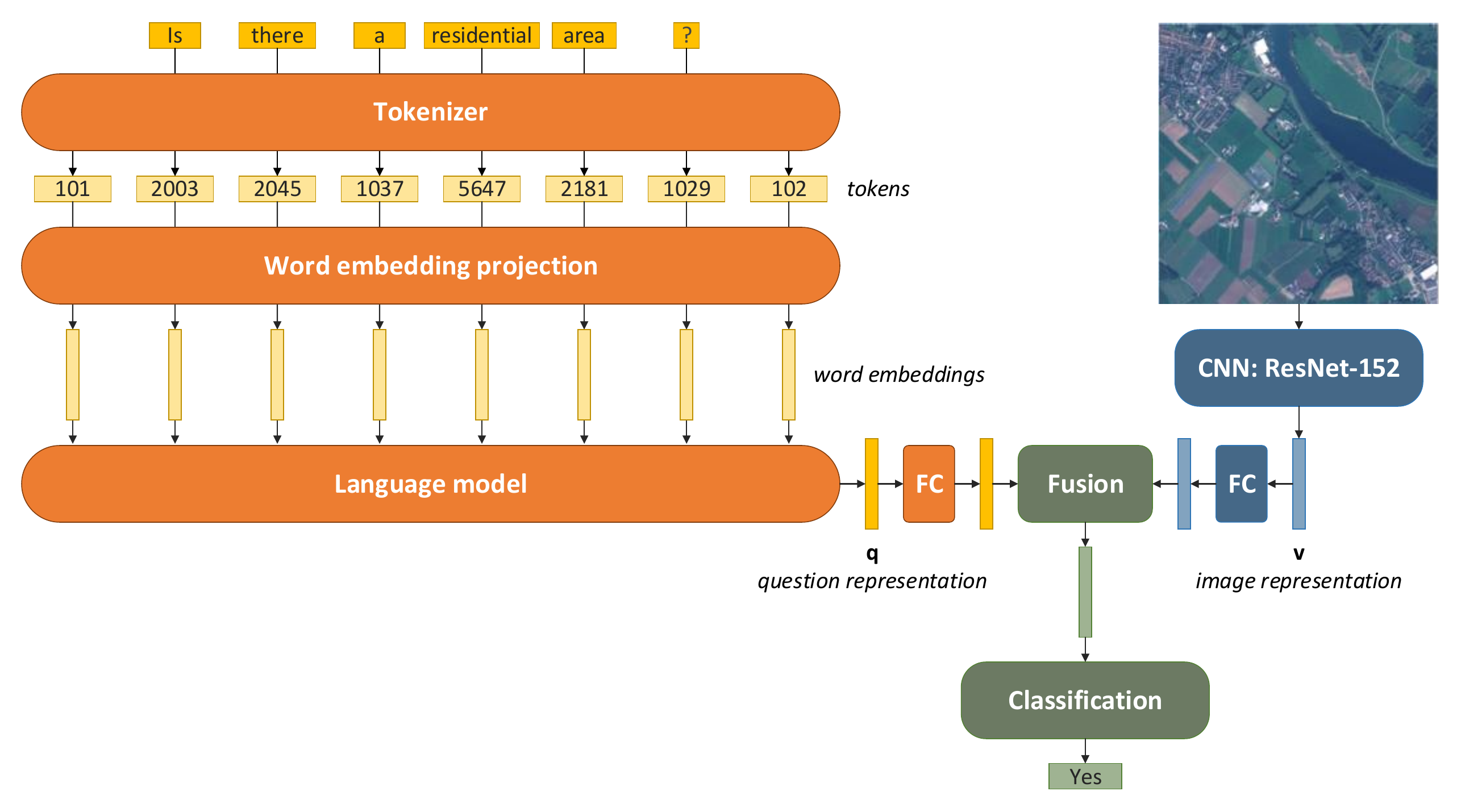}
    \caption{Architecture of a transformer-based RSVQA model~\cite{chappuis_language_2022} with a focus on the language part, composed of a tokenizer, word embedding and language model. In this work, we evaluate two such models, the Skip-thoughts~\citep{kiros_skip-thought_2015} (RNN) used in the baseline model~\citep{lobry_rsvqa_2020}, versus a Tranformer (BERT~\citep{devlin_bert_2019}, distilBERT~\citep{sanh_distilbert_2020} and roBERTa~\citep{liu_roberta_2019}).
    The element-wise multiplication used as fusion in the baseline is assessed against a concatenation of feature vectors and MUTAN~\citep{ben-younes_mutan_2017}.}
    \label{fig:Model}
\end{figure}

\section{Language bias analysis methods} 
\label{sec:methods}

We first describe the architecture of RSVQA in sub-section~\ref{subsec:methodRSVQA}. This architecture serves as a framework for our analytical study, as well as the language models. We include a discussion on the training setup. 
The three other sub-sections describe our analysis strategy, which serves the main contribution of this work. Sub-section~\ref{subsec:methodVBlind} presents the ``visual blind'' model, where only the question is considered as input, without the image; Sub-section~\ref{subsec:methodAdVal} explains the process of adversarial testing used on the full architecture.
Finally, Sub-section~\ref{subsec:protocol} describes the tailored strategy to manually analyze the datasets and practically understand the short-cuts between questions and answers a model can make for predictions.

\subsection{RSVQA model}
\label{subsec:methodRSVQA}
The general RSVQA architecture used in this work is presented in Figure~\ref{fig:Model}. We investigate and compare two different types of language models without or with fine-tuning, RNN and Transformers, each one providing the representation summarizing the question. 

Language models have received considerable attention in recent NLP and computer vision research, and witnessed in the last decade an important evolution in their architecture and achievements. In particular, while the recurrent neural network was predominant in NLP, we observe a shift towards Transformer~\citep{vaswani_attention_2017} architectures in the last years' researches. We compare a method from both categories in this work.

Practically, language models start by transforming the words of the input sentence into tokens. Tokenizers assign a unique value to each word, indexing the position of the word in the tokenizer, similarly to a dictionary. The sentence, as a sequence of tokens, is then fed to the language model that first proceeds to the extraction of word embeddings. Each token is mapped to a vector using specific methods similar to Word2Vec~\citep{mikolov_efficient_2013}, and these vectors are finally processed as inputs by the RNN or Transformer. At the end of the process, the entire sentence is converted into a vector representation $\mathbf{q}$ of dimension $\mathrm{d_q}$.

\paragraph{RNN Language model: Skip-thoughts} 

\begin{figure}[h]
    \centering
    \includegraphics[width =1\textwidth]{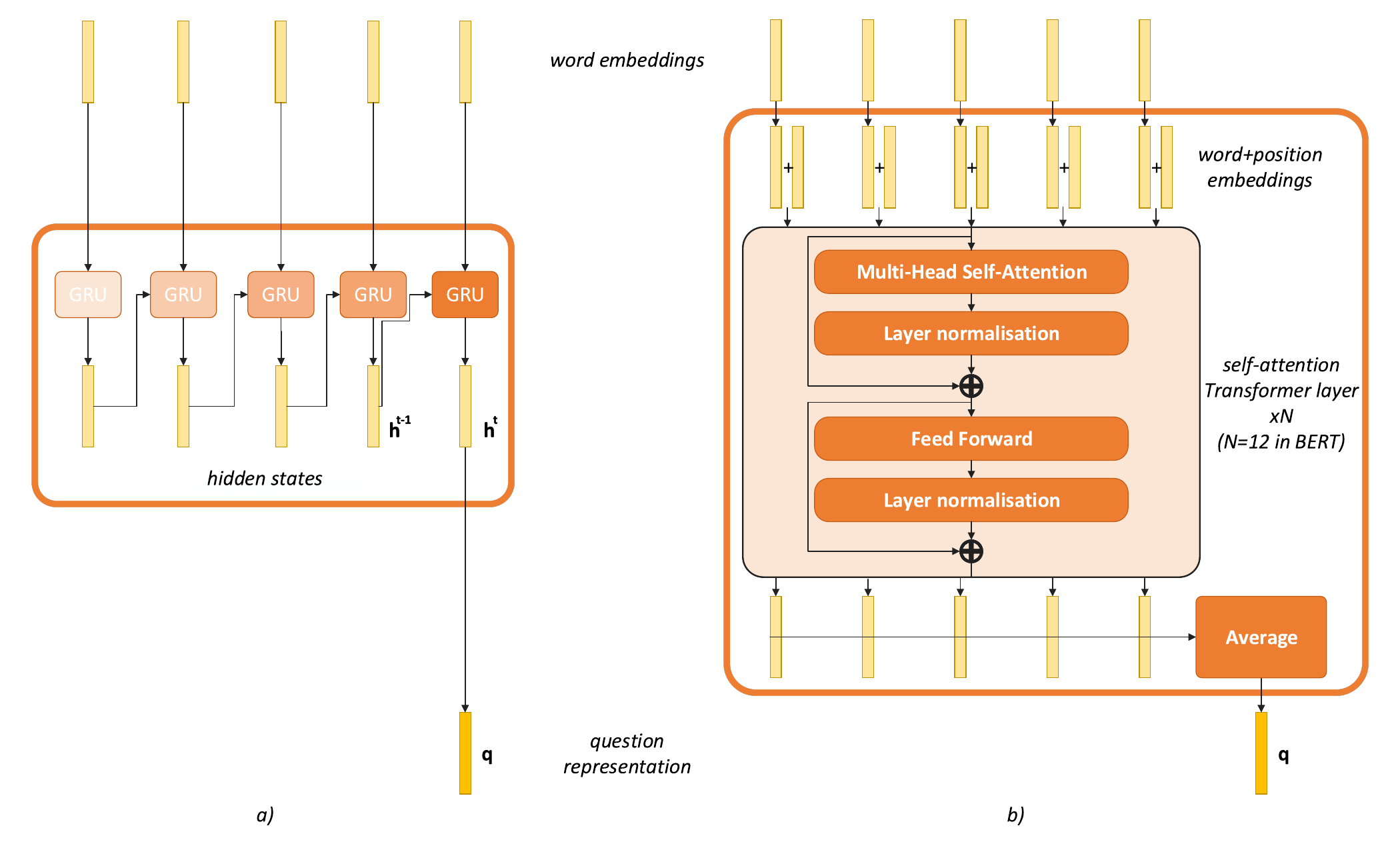}
    \caption{Details on the architecture of the two types of language models used in the RSVQA baseline for this analytical study. a) In the Recurrent Neural Network (RNN), a single GRU visits the different positions in the order, to extract meaningful information from word representations and produce a sequence representation. b) A Transformer architecture is also tasked to produce a meaningful representation of the sequence. However, instead of processing the input as a sequence, it enables the interaction between all positions directly. In the Transformer layer, residual connections are used and represented with a $\bigoplus$ sign}
    \label{fig:LM}
\end{figure}

Skip-thoughts~\citep{kiros_skip-thought_2015} is a recurrent neural network processing the text input as a temporal sequence. Going through each position of the sequence in order, a RNN takes the previous step as input to the computation of the current step, as illustrated in Figure~\ref{fig:LM} (a) for a sequence classification. Skip-thoughts is made of a Gated Recurrent Unit (GRU)~\citep{cho_learning_2014}, which is better equipped to cope with a certain tendency of RNN to forget the first positions of the sequence as it reaches its end. 
The GRU cell encodes the $620$-dimensions embedding to a hidden representation in a $\mathrm{d_q}=2'400$-dimensional space. 
In RSVQA, the last hidden state of the sequence produced by Skip-thoughts becomes the output representation $\mathbf{q}$.

As tokenizer for the Skip-thoughts in the RSVQA method, a dictionary of all words contained in the dataset is produced. A shortcoming is the impossibility to embed words outside of this specific dictionary, although the language model itself has been trained on a much larger corpus. 

\paragraph{Transformer language model: BERTs}

The Bidirectional Encoder Representations from Transformers~\citep{devlin_bert_2019} (BERT) is an attention-based language model that achieved, at the time of its publication, state-of-the-art performances on eleven NLP tasks. The Transformer architecture stacks several layers of self-attention (12 layers in standard BERT) as originally proposed by~\citet{vaswani_attention_2017}. While a RNN processes the sequence in order, limiting long-distance dependencies, even with memory mechanisms, with the Transformer attention mechanisms all positions attend to all positions in the previous layer. 
Thanks to unconstrained attention mechanisms in BERT, the input can be attended bidirectionally, taking both left and right context into consideration rather than considering only a forward reading of the sequence. 

Other models derive from BERT, among which two are considered in this study. distilBERT~\citep{sanh_distilbert_2020} is a lighter version (6 attention layers) trained using knowledge distillation~\citep{hinton_distilling_2015} that could retain 97\% of BERT base performance on GLUE benchmark.
roBERTa~\citep{liu_roberta_2019} is a model with the same architecture as the BERT large model (24 attention layers), but pre-trained for a longer period of time on more data, over 160GB of uncompressed text against about 16GB for BERT. 

Technically, in a Transformer architecture, to enforce the notion of positions, a positional embedding is added to the initial word embedding input, as illustrated in Figure~\ref{fig:LM} (b).
In each self-attention Transformer layer, the vectors go first through a step of multi-head self-attention where attention scores are computed and used similarly to a weighted average of the previous layer.
As the attention mechanism is a linear operation, 
a fully-connected layer is applied after the attention layer and is referred to as a feed-forward layer. Its weights are learned during training and the same parameters are applied to all positions. 
To improve the training of a Transformer, residual connection and normalization are used after both multi-head self-attention and feed forward operations.

In this work, we compared the performance of three BERT models: BERT base, distilBERT~\citep{sanh_distilbert_2020} and roBERTa~\citep{liu_roberta_2019}, on a validation round before selecting the best performing for further experimentation and analysis. 
The original BERT and distilBERT use a WordPiece tokenizer~\citep{wu_googles_2016} with a vocabulary size of $30'522$. 
Going further, roBERTa tokenizer (Byte-Pair Encoding) represents text at a hybrid level between characters and words, enabling it to handle an even larger vocabulary.
Not limited to the words of the dataset, these tokenizers could directly transfer to another dataset without any additional pre-processing. 
Tokens are embedded with their position in a space of $\mathrm{d_q}=768$ dimensions by the word embedding.
We use the transformers package from Huggingface~\citep{wolf_huggingfaces_2020}: this open library provides an implementation of both specific tokenizers and language models.
For BERT, distilBERT and roBERTa, the pre-trained weights are retrieved from the main checkpoint ``bert-base-uncased'', ``distilbert-base-uncased'' and ``roberta-base'' respectively.
In this work, an average across all positions of the last hidden layer is employed as the output representation $\mathbf{q}$.


\paragraph{Visual encoder} In parallel to the language part described above, visual information is encoded with a Convolutional Neural Network, a typical type of architecture for the visual input (see the right side of Figure~\ref{fig:Model}). The visual feature extractor is a ResNet-152~\citep{he_deep_2016} whose weights are pre-trained on Image-Net~\citep{deng_imagenet_2009} and kept frozen in this study. The output dimension of this encoder is $2'048$. 

\paragraph{Multimodal fusion}
After independent extraction, both visual and textual features, whose dimension $\mathrm{d_q}$ differs depending on the language model ($\mathrm{d_q = 2400}$ for Skip-thoughts and  $\mathrm{d_q=768}$ for BERT models), are projected in a space of $1'200$ dimensions each. 
Inspired by~\citet{chappuis_how_2021}, different strategies to combine the language and visual representations in the fusion step (element-wise multiplication, MUTAN~\citep{ben-younes_mutan_2017} and concatenation) are tested to assess the influence on the performances combined with different language models. Element-wise multiplication is the easiest fusion strategy where each position of image and text vectors are multiplied together ($f=v \odot q$, also referred to as Hadamard product), to create a new vector of the same shape ($\mathrm{d_f=1200}$ in this case). In the concatenation, vectors are stacked, $f=[v;q]$. In the MUTAN strategy, the fusion is richer and uses a Tucker decomposition, and is learned during training (details in~\citet{chappuis_how_2021}).

\paragraph{Classification}
Finally, the product of the fusion step, f, is classified into answers with a fully-connected layer of size $256$ before being projected to the answer space (bottom part of Figure~\ref{fig:Model}). The last layer of the model has a size as large as the number of possible answers in the dataset. Designing the task as a classification simplifies it, while limiting it (i.e. the model cannot predict answers other than pre-defined ones). The answer with the highest activation is selected as the model prediction given the image and question inputs.

\paragraph{Model training setup}
For each language model, we perform two experiments: one for methods with frozen language encoder weights (i.e. the pre-trained language model encoder, without fine-tuning) and a second, where the language model is fine-tuned alongside the rest of the RSVQA fusion and classification layers. By fine-tuning, we hope that the language model will close a potential domain gap from the pre-trained encoders to the task-specific and domain-specific language used.
A validation round is performed with all language Transformers and fusion strategies. The best method is selected and run on the test set. 

For all experiments, the batch size is 70, the length of the question embedding is 50, Adam~\citep{kingma_adam_2015} is used as optimizer and models are trained during 50 epochs. The learning rate of the fusion and classification layers of the model is ${10^{-5}}$. Different fine-tuning learning rates have been selected for Skip-thoughts and BERTs, ${10^{-4}}$ and ${10^{-6}}$ respectively, following a hyper-parameter search on the validation set.
Three runs of experiments are conducted for each selected model and we report the average performance and its standard deviation.

\subsection{Language bias analysis: Visual blind model}
\label{subsec:methodVBlind}
\begin{figure}[!t]
    \centering
    \includegraphics[width =0.8\textwidth]{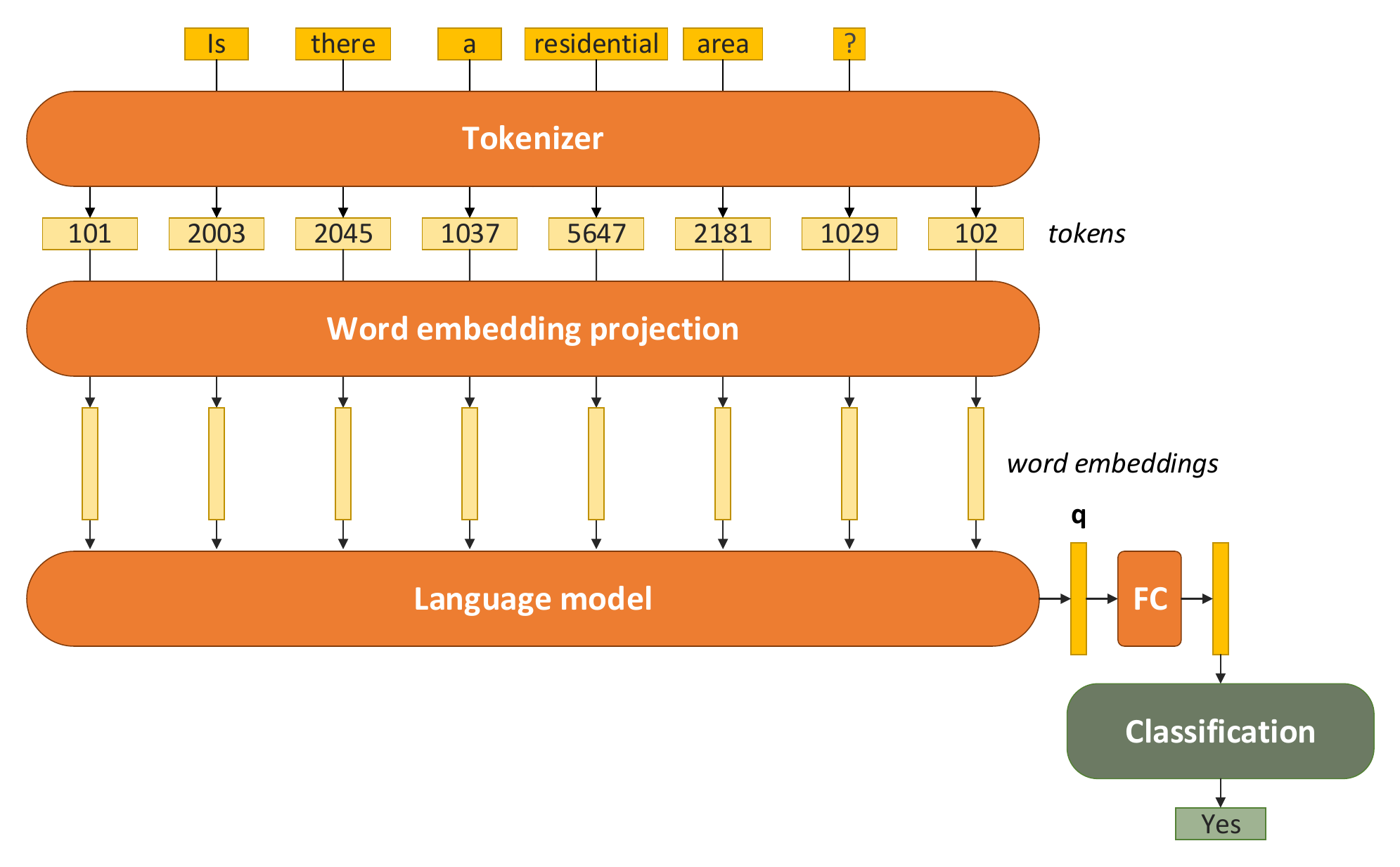}
    \caption{In the ``visual blind'' architecture, the question representation is directly fed into classification, without any combination or information from the image representation}
    \label{fig:BlindModel}
\end{figure}

In order to identify language biases in RSVQA, ``visual blind'' models are implemented and assess how much can be learned from the textual information alone, by a deep learning architecture. As illustrated in Figure~\ref{fig:BlindModel}, the only input to the model is the textual question, and its representation, extracted by the language model, is directly fed to the classification layers of the network. Hence, these models never see the image and rely exclusively on the information present in the question, leveraging the distribution of question/answer in the dataset and highlighting language biases.

The ``visual blind'' model is trained with the same hyper-parameters (including number of epochs) as the reference model. Results with the language model frozen or fine-tuned alongside training of the classification layers are compared.

\subsection{Language bias analysis: Adversarial testing}
\label{subsec:methodAdVal}
Contrary to visual blind models that predict an answer based only on the question, adversarial testing shows the importance of visual information for the answer prediction with a reference model trained on both question and image.
Indeed, while it might be difficult to avoid any leveraging of language biases by RSVQA models, showing that these models depend on the visual information as well is beneficial to determine their robustness.

To illustrate the importance of the visual information for fully-trained model and compare it to visual blind models, we use a process referred to as adversarial testing. Run only at inference, the trained RSVQA models are provided with the sample question, targeting the sample answer, but with a tampered image. To avoid time-consuming manual modifications of the images triggering different answers, the image can be whited- or blacked-out, or can be changed at random from the entire dataset. While whited- or blacked-out images represent a full disturbance of the visual field, they do not belong to the typical profile of images in the dataset. Selecting an image at random in the dataset on the other hand has the advantage of remaining in the same visual distribution and thus appears as a more realistic and meaningful strategy. Additionally, for a few samples, we manually manipulated part of a few images with a photo-editing software, with the goal to trigger the prediction of a different answer from the models. While this strategy is not feasible on a large scale, it allows to illustrate a source of language bias typical to remote sensing: geographical spatial similarities.

\subsection{Language bias analysis: Datasets analysis}
\label{subsec:protocol}
To identify biases the language models may learn, we perform an analysis of the dataset. The strategy is to define subsets of the dataset, at a higher-level by question types, or at a lower-level by the objects and relations categorisation in the question. We then compare the performance of the RSVQA models against the one of 1) a system identifying the most common answer, or prior, for each subset and 2) a completely balanced system with identical probability for all possible answers.

\paragraph{By question types}
Questions are often grouped into different categories, usually relative to their main task (i.e. detecting the presence, counting objects, etc) or level of difficulty. Additionally, the number of possible answers is also limited in the closed-world environment of a dataset. We compare two settings in which answers are based on question type only:
1. Going through the training set, the most common answer is defined for each question type. This most common answer is assigned as prediction to all samples of this question type in the testing split (when available) and an accuracy is calculated by comparing these predictions to the answer targets.
2. The uniform accuracy, which corresponds to the result obtained with a uniform answer distribution, if all answers within each question type had the same probability. 

Should there be no clearly assigned question types for the samples of a dataset (i.e. RSIVQA), image-question-answer triplets can be grouped by their type of answers, e.g. numerical, yes/no (binary) and text. For the datasets examined in this study, question types used are detailed in the dataset descriptions.

\paragraph{By objects and relations}
Digging deeper, the question construction scheme of the dataset is investigated. Questions can often be dissected and grouped as the combination of objects, relations and attributes. In this study, this deeper analysis is only performed on the RSVQA Low resolution dataset~\citep{lobry_rsvqa_2020}, presented in more details in Section~\ref{sec:DA}. However, the same principle can be applied to other datasets as well. Since RSVQA Low resolution was constructed by following the CLEVR protocol~\citep{johnson_clevr_2017} on OpenStreetMap data, we have access to the relational tree used to build the image-question-answer triplet, therefore we can straightforwardly analyse biases at the level of object and relation. Each question can be deconstructed and associated to subsets. As for the analysis of the question types above, we proceed by considering both the most common and uniform accuracy strategies.
In this work, when performing the finer analysis, the questions are grouped based on the objects of interest (e.g. roads, buildings, water areas, etc.), with or without attributes.

\section{New evaluation metrics accounting for biases}
\label{sec:evalMetrics}
In RSVQA research, and more generally when casting VQA as a classification problem, the common evaluation metric is an accuracy measure. It can be an overall accuracy considering all samples, or a specific accuracy on defined fractions of the dataset, for example calculated by question types. While this metric is common and relatively simple to understand, it has the disadvantages of hiding the problem of language biases and giving the deceiving impression of high performances. Since RSVQA is formulated as a classification method, an answer will inevitably be provided, and therefore a model basing its answers only on language biases (e.g. always providing the most common answer for a given question) could reach good performances for the wrong reasons. 

While accuracy remains the usual choice for measuring performance, we propose here a relative evaluation strategy and in particular two metrics taking language bias into account. In one case we focus only on the dataset, and in the second case we consider both the dataset and the model. 
The relative evaluation considers a lower-bound of the accuracy range to highlight the improvement of a RSVQA model. We propose to report this improvement as a normalized, complementary evaluation metric, alongside the typical overall accuracy metric. To allow comparison, improvement metrics are calculated on the test set of the dataset.
The lower-bound of accuracy is defined differently in the two metrics and will be detailed below. 
On the other end of the spectrum, the best performance, the upper-bound, corresponds to 100\%. 
Considering the accuracy of a RSVQA specific model, $\mathbf{RSVQA\_Acc}$, and the lower-bound of accuracy, $\mathbf{LowerBound\_Acc}$, we calculate the improvement of accuracy over the lower-bound, $\mathbf{IO\_LowerBound}$, as:

\begin{equation}
\label{IO_LB}
    \mathbf{IO\_LowerBound} = \frac{\mathbf{RSVQA\_Acc} - \mathbf{LowerBound\_Acc}}{1 - \mathbf{LowerBound\_Acc}}
\end{equation}

\begin{itemize}
    \item Firstly, in the ``dataset-only lower-bound'', we propose to calculate the minimal performance by using the probability of the most common answer in specific fractions of the dataset (see Section~\ref{subsec:protocol}). This can be calculated at the level of the entire dataset, question types, or more detailed levels, and represents the minimal performance we expect a model should achieve. In particular, identifying question types should be a simple task for a language model. 
    Let $|\mathcal{A}_s|$ be the number of possible answers for the dataset subset $s$ (e.g. counting questions), 
    $N_s$ the number of samples in subset $s$, 
    and $n_s^{prior}$ the number of questions with the most common answer, 
    the different accuracy measures are defined as $\mathbf{Prior\_Acc}_{s}$ for the ``most common answer'' or prior measure, and $\mathbf{Uni\_Acc}_{s}$ for the uniform measure. 

    \begin{equation}
        \mathbf{Prior\_Acc}_{s} = \frac{n_s^{prior}}{N_s} \\
    \end{equation}
    \begin{equation}        
        \mathbf{Uni\_Acc}_{s} = \frac{1}{|\mathcal{A}_s|} \\
    \end{equation}

    The improvement over the lower-bound of accuracy in this ``dataset-only'' setup, $\mathbf{IO\_Prior}$, is calculated with $\mathbf{Prior\_Acc}$ as $\mathbf{LowerBound\_Acc}$ in \autoref{IO_LB}. This ``dataset-only'' measure relates exclusively to a classification setup where the model selects one class among a set of possible answers (i.e. as RSVQA is framed in existing work~\citep{lobry_rsvqa_2020, yuan_easy_2022,felix_cross-modal_2021,bazi_bi-modal_2022,chappuis_prompt-rsvqa_2022}). 

    \item Secondly, we derive an evaluation metric considering both dataset and model, by considering a model-based lower-bound of accuracy, as $\mathbf{LowerBound\_Acc}$. 
    Selecting the adversarial testing, the motivation is to show the improvement of accuracy for a reference model provided either with the appropriate image of the sample or a corrupted image. This improvement measure highlights the predictions dependant on the visual input.
    The results obtained with the adversarial testing use the actual RSVQA model and only require inference, without any re-training. 
    This second relative evaluation metric is referred to as $\mathbf{IO\_AdTest}$ and is calculated with $\mathbf{AdTest\_Acc}$ as $\mathbf{LowerBound\_Acc}$ in \autoref{IO_LB}.
    Alternatively, visual blind models are also interesting candidates. However, they would imply a certain computation cost for the evaluation itself, as these systems are trained similarly to the actual model.

\end{itemize}

\section{Datasets}
\label{sec:DA}
\label{subsec:datasets}
\begin{figure}[!t]
    \centering
    \includegraphics[width =\textwidth]{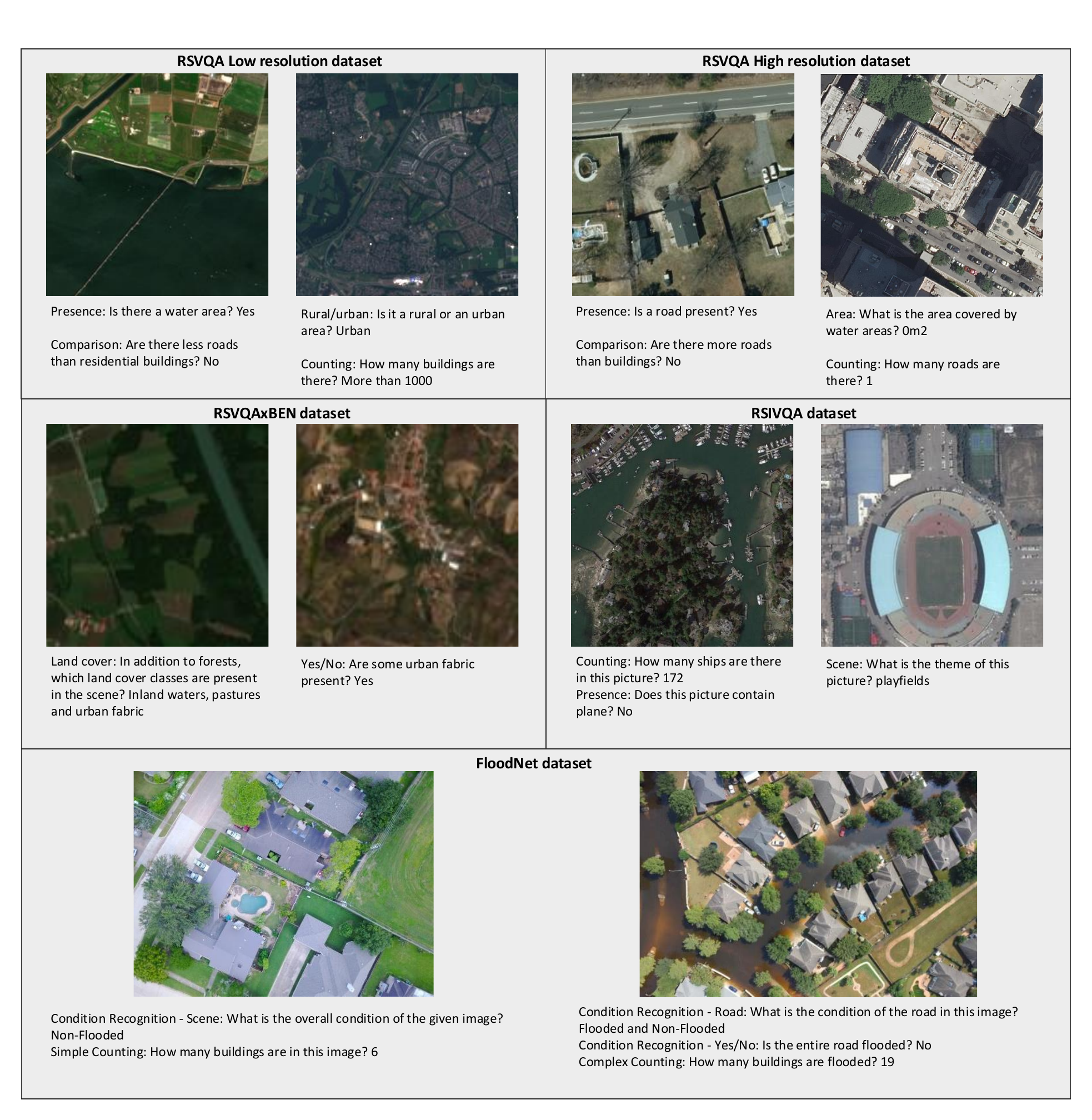}
    \caption{Samples of image/question/answer triplets for each dataset looked at in this study. The question types are specified as well.}
    \label{fig:Samples}
\end{figure}

Advocating for a better understanding of the data used to train models, we describe in Section~\ref{subsec:datasets} the datasets available for the task of RSVQA (with question/answers for single images). A few samples of these datasets are displayed in Figure~\ref{fig:Samples}.


\paragraph{RSVQA - Low resolution~\citep{lobry_rsvqa_2020}} Model comparisons are conducted on the low resolution set of the RSVQA dataset\footnote{Available at \href{http://rsvqa.sylvainlobry.com}{rsvqa.sylvainlobry.com}}, composed of 772 Sentinel-2 RGB images (of size $256\times 256$ pixels with a 10m resolution) over the Netherlands and 77'232 image/question/answer triplets. Questions and answers are automatically generated using OpenStreetMap vector layers and with a generation strategy inspired by the CLEVR protocol~\citep{johnson_clevr_2017}. 
In the case of this dataset, there are four categories: Comparison, Counting, Presence, Rural/urban. A total of nine answer classes are present in RSVQA Low resolution: Yes, No, 0, between 1 and 10, between 11 and 100, between 101 and 1000 and more than 1000, rural, urban. 

RSVQA Low resolution is used to run the experiments with varying language models and fusion strategies in the baseline RSVQA architecture, and to fully exemplify the two different levels of dataset analysis described in Section~\ref{subsec:protocol}.

\paragraph{RSVQA - High resolution~\citep{lobry_rsvqa_2020}}
This dataset consists of aerial imagery from the USGS collection (15cm spatial resolution) for a total of 1'066'316 image-question-answer triplets (train, validation and test sets included). The geographical locations covered are four regions in the North East Coast of the USA: New York City (NY), Long Island (NY), Portland (ME) and Philadelphia (PA). There are two test sets, which differ in geographical location. Test 1 is located in New York City, a familiar area for the model as it can see images of the same area in training. Test 2 covers a zone in Philadephia, which is absent from training data.
The questions are grouped into 4 categories: Area, Comparison, Counting, Presence, and a total of 55 answer classes are present in the train set (see~\citet{lobry_rsvqa_2020}). 
Due to its higher resolution, this dataset allows to enquire about relatively small object's coverage and counting. Especially for the latter, the answers are numeric (instead of categorical ranges in RSVQA Low resolution).


\paragraph{RSVQA x BEN~\citep{lobry_rsvqa_2021}}
RSVQAxBEN dataset is a large-scale RSVQA dataset generated from 10m-resolution RGB images of Sentinel-2 (similarly to RSVQA Low resolution) and the labels from BigEarthNet~\citep{sumbul_bigearthnet_2019}, a land cover classification dataset based on Corine Land Cover. 
The dataset is organized around two types of questions: Yes/No or Land cover questions. While there is a total of 28'049 possible answers, in practice it was limited to the 1'000 most frequent answers in three published architectures~\citep{lobry_rsvqa_2021, chappuis_prompt-rsvqa_2022, siebert_multi-modal_2022}.
The split between training, validation and testing sets is separated by the latitude coordinate of the images (which is different from the random split published with BigEarthNet~\citep{sumbul_bigearthnet_2019}), making the land cover classification exercise more challenging as the distribution of classes differs in geographically distant areas of Europe. 

\paragraph{RSIVQA~\citep{zheng_mutual_2021}}
This dataset is generated from existing remote sensing datasets of image classification (UCM~\citep{yang_bag--visual-words_2010}, Sydney~\citep{zhang_saliency-guided_2015}, AID~\citep{xia_aid_2017}) and object detection (HRRSD~\citep{zhang_hierarchical_2019}, DOTA~\citep{xia_dota_2018}). In total, including training, validation and testing set, the dataset is composed of 111'693 image-question-answer triplets. The samples can be organized into 4 question types: Count, Location, Presence (Yes/No), Scene (classification). There are 92 unique questions and 579 unique answers. 
One main disadvantage is the lack of published datasets splits. The paper presenting RSIVQA suggested a random 80\% training, 10\% validation, 10\% testing splitting strategy. However, without clearly making this distribution available to the research community, new experiments (defining similar distribution but with different samples due to randomness) are difficult to compare. With respect to the language bias analysis in Section~\ref{subsec:DAresults_extension}, instead of randomly splitting the dataset, the dataset analysis is performed on the entire dataset at once.

\paragraph{FloodNet~\citep{rahnemoonfar_floodnet_2020}}
This high resolution dataset focuses on the thematic of natural disasters. It contains images from an unmanned aerial vehicle (UAV) after the passage of hurricane Harvey in the USA in August 2017. The data acquired after this event are used for two tasks: image segmentation and visual question answering. Regarding the RSVQA task, the samples are organized into 3 questions types: Condition Recognition, Simple Counting, Complex Counting. While Simple Counting questions query the frequency of an object in the image (e.g. buildings), Complex Counting questions also consider an attribute of the object (e.g. flooded / non-flooded buildings). Condition Recognition questions consist of three easily defined groups (questions answered with yes/no, questions about the condition of the road, and the overall condition of the scene) and thus lead to the subdivision of this question type into: Condition Recognition - Yes/no, Condition Recognition - Road, Condition Recognition - Scene. 
The dataset consists of only 15 different phrasing of questions. 
There are a total of 41 answer classes, divided between 5 textual answers and the rest as numerical answers.
Unfortunately, only the training set is publicly available in full, i.e. with answers. Thus, the dataset analysis in Section~\ref{subsec:DAresults_extension} is only performed on the training set. 

\section{Results and discussion}
\label{sec:results}
\subsection{Language transformers in remote sensing visual question answering}
\begin{table}[!t]
\caption{Validation results on RSVQA Low resolution, by overall accuracy. LM refers to language model and frozen indicates that the weights of the language model are kept as pre-trained and remain unchanged during the training process. In terms of fusion strategy, ``e-mult'' refers to element-wise multiplication and ``concat'' for concatenation. At this step, one of the three Transformers as well as a fusion is selected and highlighted in bold.
}
\label{tab:ValresLR}
\centering
\resizebox{\linewidth}{!}{%
\begin{tabular}{l|c|c|c|c|c|c|c|c|c|c|c|c|}
\hline
& \multicolumn{3}{c|}{RNN} & \multicolumn{9}{c|}{Transformer} \\
\hhline{~~~~~~~~~~~~~}
& \multicolumn{3}{c|}{Skip-thoughts (RNN)} & \multicolumn{3}{c|}{BERT} & \multicolumn{3}{c|}{\textbf{distilBERT}} & \multicolumn{3}{c|}{roBERTa} \\
\hhline{~~~~~~~~~~~~~}
Fusion strategy & e-mult & concat & \textbf{MUTAN} & e-mult & concat & MUTAN & e-mult & concat & \textbf{MUTAN} & e-mult & concat & MUTAN \\
\hline
\textbf{LM frozen} & 80.17 & 79.60 & \textbf{82.97} & 73.17 & 72.90 & 74.93 & 73.40 & 72.70 & \textbf{75.20} & 71.20 & 70.70 & 72.10 \\
\hline
& \multicolumn{3}{c}{} & \multicolumn{3}{|c}{} & \multicolumn{3}{|c}{} & \multicolumn{3}{|c|}{} \\
& \textbf{e-mult} & concat & MUTAN & e-mult & concat & MUTAN & \textbf{e-mult} & concat & MUTAN & e-mult & concat & MUTAN \\
\hline
\textbf{LM fine-tuned} & \textbf{87.43} & 86.70 & 87.07 & 87.27 & 87.03 & 86.93 & \textbf{87.90} & 86.97 & 87.43 & 87.60 & 86.90 & 87.23 \\
\hline
\end{tabular}}
\end{table}

Based on validation results in Table~\ref{tab:ValresLR}, a selection is made among the three language Transformers and three fusion strategies. For Transformers, the best performances on validation are obtained with distilBERT, which is also the lightest of the three. It is selected for all following work. This observation and choice tend to show that the language diversity and difficulty of current remote sensing VQA datasets is well captured by the smaller size of Transformer and does not benefit from larger capacities.
For the fusion strategies, MUTAN~\citep{ben-younes_mutan_2017} performs better when the language models weights are frozen. It corresponds to findings in \citet{chappuis_how_2021}. However and interestingly, when fine-tuning the language model, the element-wise multiplication appears as the most performing fusion strategy. 

Following this validation step, the selected options are run on the test set and methods with frozen or fine-tuned language model are compared, in addition to the comparison between the RNN and Transformer NLP architecture. The results are reported in Table~\ref{tab:resultsLR}. They are detailed by question type: Comparison, Counting, Presence, Rural/urban.

\begin{table}[!b]
\caption{Numerical results on the RSVQA, low resolution, task. AA and OA represent average accuracy and overall accuracy, respectively. Results are averaged over three runs, with standard deviation in brackets. 
}
\label{tab:resultsLR}
\centering
\resizebox{0.99\linewidth}{!}{%
\begin{tabular}{l|l|l|c|c|c|c|c|c|c|}
\hhline{~~~-------}
\multicolumn{3}{c|}{} & \multicolumn{7}{c|}{Accuracy}\\
Language model & LM weights & fusion & Comparison & Counting & Presence & Rural/urban & AA & OA\\
\hline
RSVQA baseline~\citep{lobry_rsvqa_2020} & frozen & e-mult & 81.50 (0.03) & 67.01 (0.59) & 87.46 (0.06) & \textbf{90.00 (1.41)} & 81.49 (0.49) & 79.08 (0.20)\\
\hline
\hline
RNN Skip-thoughts & \multirow{2}{*}{frozen} & \multirow{2}{*}{MUTAN} & 85.47 (0.23) & 66.88 (0.64) & 89.02 (0.08) & 86.67 (0.47) & 82.01 (0.16) & 81.05 (0.21)\\
Transformer distilBERT & & & 68.26 (0.21) & 64.70 (0.69) & 88.58 (0.25) & 89.33 (2.05) & 77.72 (0.29) & 73.42 (0.27)\\
\hline
\hline
RNN Skip-thoughts & \multirow{2}{*}{fine-tuned} & \multirow{2}{*}{e-mult} & 90.59 (0.08) & \textbf{70.82 (0.46)} & 90.51 (0.06) & 89.33 (4.50) & 85.31 (1.12) & 84.73 (0.16)\\
Transformer distilBERT & & & \textbf{90.76 (0.26)} & 70.59 (0.36) & \textbf{91.04 (0.30)} & \textbf{90.00 (1.41)} & \textbf{85.60 (0.36)} & \textbf{84.90 (0.05)}\\
\hline
\end{tabular}}
\end{table}

Comparing results of the models with a frozen language model (top rows of Table~\ref{tab:resultsLR}), Skip-thoughts captures already well the language information in the system without any fine-tuning. This is likely due to the dimension $\mathrm{d_q}$ of the language encoder output that is more than 3 times larger for Skip-thoughts than distilBERT ($2'400$ and $768$, respectively). The RNN might simply retain more information useful for the task. This method however requires to train more parameters as the fully-connected layer projecting the $\mathrm{d_q}$-dimensional encoding to the next vector is much larger with the RNN than distilBERT. 

Looking at the results with the fine-tuned models in Table~\ref{tab:resultsLR} (bottom) both models perform similarly and always out-perform their frozen counterpart. In both cases, a strong increase of performance is observed, with increases in overall accuracy between 3\% (Skip-thoughts) and 10\% (distilBERT). distilBERT performs better than Skip-thoughts across all question categories, except ``Counting''. This question type is also the more challenging across the four question types. 
The less stable results, with the largest standard deviation, belong to the question type ``Rural/urban''. This can be explained by the relatively low number of samples for this category (1\% of the dataset). 
Some categories are improved by a large margin with the Transformer (e.g. comparison questions improve by 20\%), while the overall accuracy gains about 10\%. This denotes the importance of fine-tuning especially for distilBERT, to adapt to the task at hand. 
It can be argued that, in the task of RSVQA and in the simple type of architecture deployed in this study, the smaller-size output of distilBERT requires an adaptation with fine-tuning to focus on and retain the relevant information, and compensate the semantic domain gap.

Beside performances, the vocabulary capacity, i.e. the number of words ``recognized'' by the tokenizer, is vastly different between the specific tokenization implementation for Skip-thoughts in this study and the distilBERT tokenizer. The limitations of the manual dictionary for Skip-thoughts, using only words of the dataset, are critical with respect to distilBERT tokenizer, which recognizes a 30k-words vocabulary. Thus, the distilBERT tokenizer is more appropriate to the potential variability of natural language, in particular when considering generalizing to other RSVQA dataset. While a more versatile tokenizer allows for more diverse inputs, there is no doubt that framing the RSVQA task as a classification problem still limits the output of the methods to pre-defined answer classes independently of the language model used.
Regardless, this point illustrates an important discussion about the language variety relevant to satellite imagery. Although the RSVQA vocabulary is common, potential use cases could require a specific vocabulary with implications on the tokenizer and word embedding. Alternatively, while a user should be able to freely formulate requests, their variety can be questioned, in particular considering low image resolution.

\subsection{Visual blind models and adversarial testing}
\label{subsec:blindAd_discussion}
In this Section, the influence of language biases on the predictions of RSVQA models are examined in two ways.
First with visual blind models, we assess the performances of systems fed only with the question as input. In this case, visual blind models have their NLP part either left as pre-trained (their weights are frozen), or fine-tuned alongside the classification layers, for the RSVQA task. 
Second is adversarial testing, a process run at inference only, where the image is switched at random with another image of the dataset (see Section~\ref{subsec:methodAdVal}). 
All the results are illustrated in Figure~\ref{fig:Full_blind_adv} (specific numerical results are reported in Appendix, Table~\ref{tab:blindAdResultsLR}). Adversarial testing experiments with whited- or blacked-out images are performed as well, and led to similar results. Therefore, they are omitted.

\begin{figure}[!t]
    \centering
    \includegraphics[width =\textwidth]{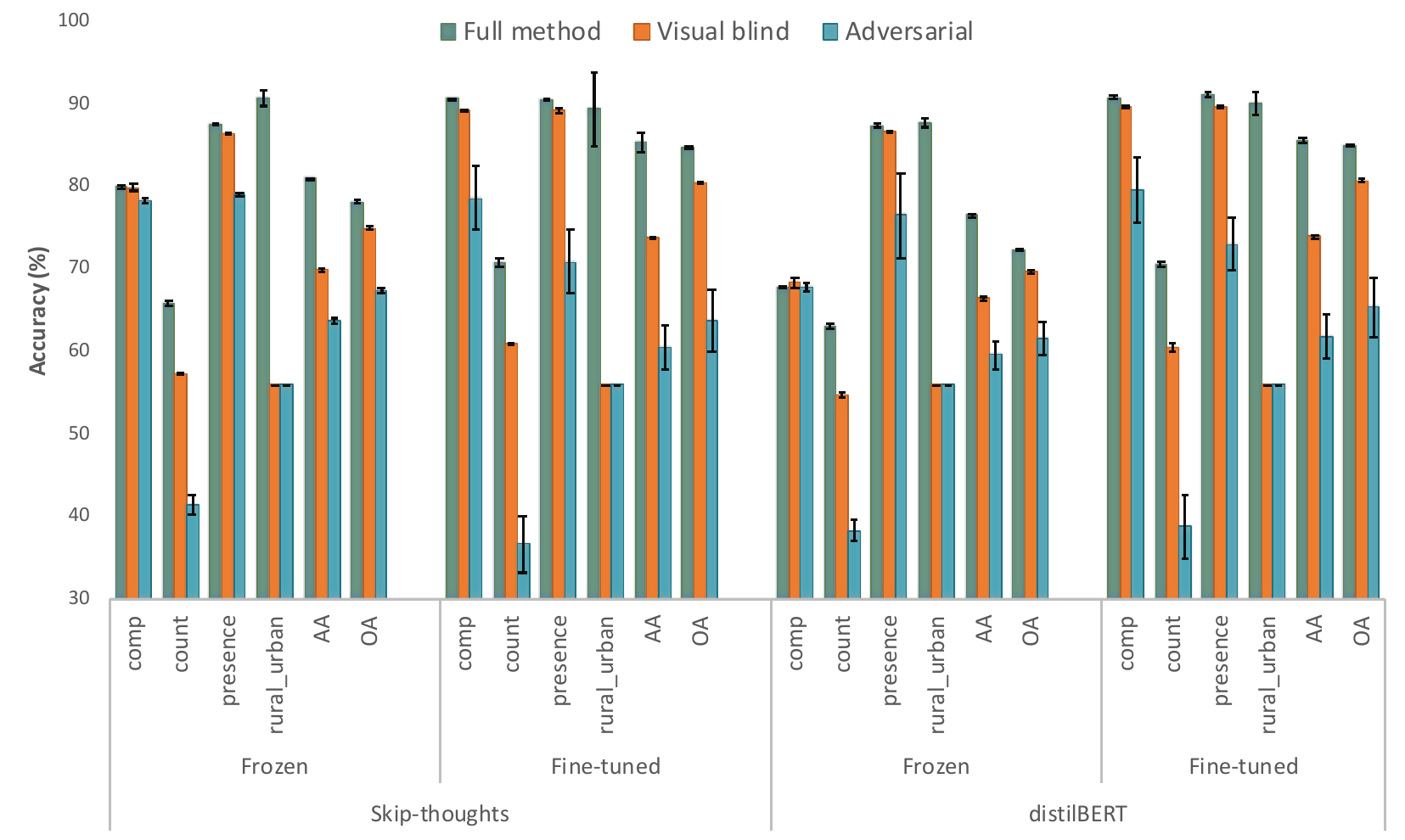}
    \caption{Testing results of the visual blind models and adversarial testing, compared to the reference models, that are provided with both question and image (averaged over three runs, with standard deviation represented by the error bars). All of these models have an element-wise fusion to remove as much additional complexity and give the focus to the assessment of language.}
    \label{fig:Full_blind_adv}
\end{figure}

Results of the reference models on the test set show performances generally above those of the visual blind models. This type of architecture thus meet the simple sanity check of reaching better performances than an equivalent structure without access to visual information.
However, results of the visual blind models are competing with the corresponding reference model for question types ``Comparison'' and ``Presence'', indicating a strong bias in their question/answer distribution. Given the question only, models are able to guess the correct answer with a good accuracy. 
Practically, it means that while the total number of ``yes'' and ``no'' answers might be roughly similar over the entire dataset, the model is able to identify elements in the questions that relates more strongly to either ``yes'' or ``no''.
Inversely, for ``Rural/urban'' questions, the accuracy of visual blind models is close to 50\%. This is explained by the question being exactly the same for all the samples of this category and an almost equal distribution of the two answers. The lack of language diversity in ``Rural/urban'' questions ensures that the language model cannot learn a more detailed relation between question and answer. However, the proportion of this question type is low in the dataset (about 1\% of all samples), thus not influencing substantially the overall accuracy. 
Comparing between full and visual blind models, with frozen or fine-tuned language models, allows to observe the effect of fine-tuning in the context of language bias.
With both language models, visual blinds are closer to the reference model when frozen than they are when fine-tuned, especially in ``Counting'' category, and to a lesser extent in ``Comparison'' and ``Presence'' questions. This tends to show that fine-tuning the language model might have helped aligning visual and textual information, leading to greater advantages over the visual blind models. 
Nevertheless, results of the visual blind models globally support the concerns of strong language biases in the dataset. Thus, there is a need to better understand the importance of adequate visual information in the answer prediction, and the adversarial testing provides a starting point in this analysis. 

\begin{figure}[!t]
    \centering
    \includegraphics[width =0.7\textwidth]{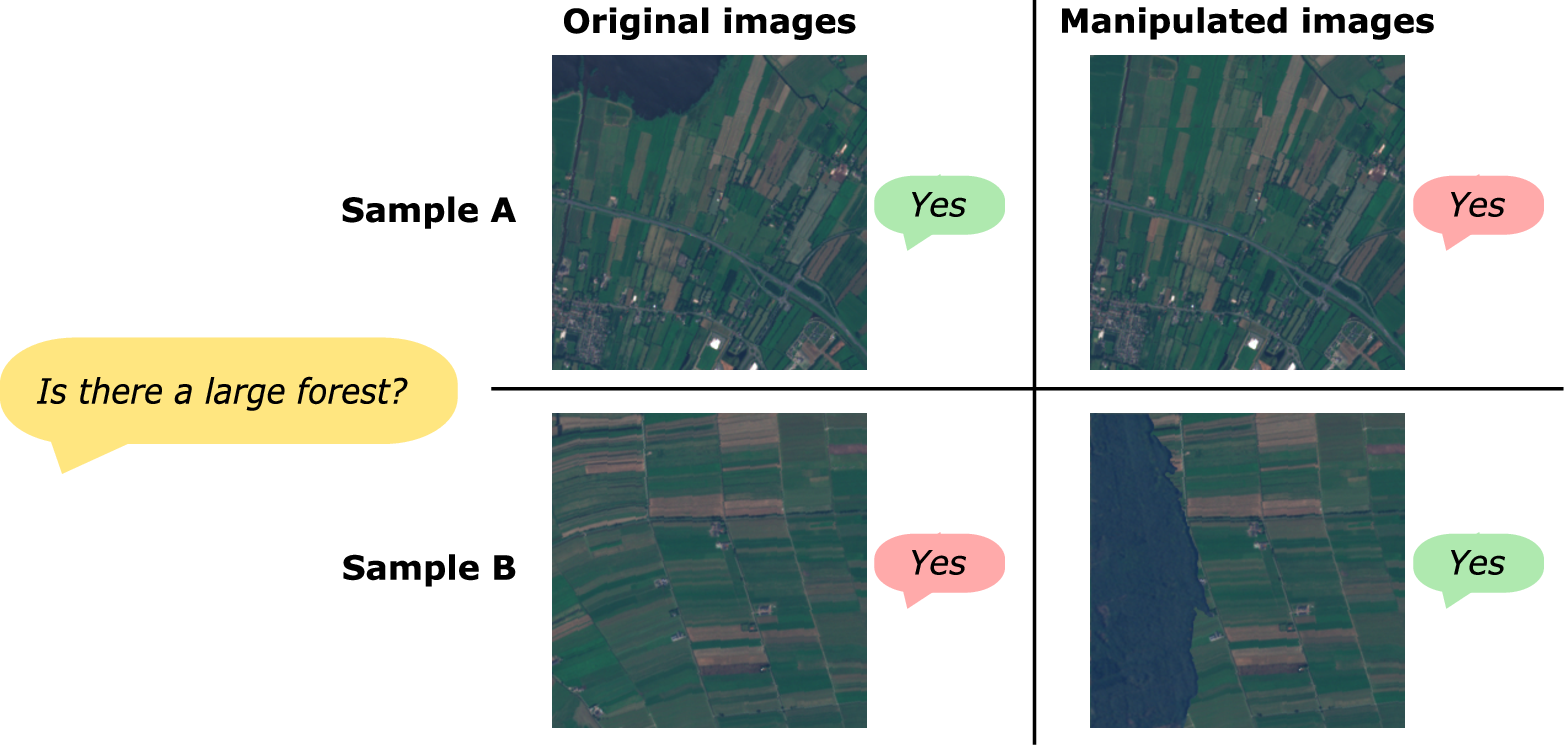}
    \caption{Adversarial testing examples. We manipulated images by removing or adding a forest patch, expecting a change in the answer to the question. Unfortunately, identical answers are predicted by all architectures experimented with in this study, regardless of the language model, supporting the issue of language bias in RSVQA. Green coloring of the answer text depicts a correct prediction, while red is incorrect.}
    \label{fig:ManualAdTest}
\end{figure}

Looking at the results of adversarial testing, it appears that providing an incorrect image to a reference model leads to worse accuracy than training a model with only the question. This tendency is even stronger in models with fine-tuned language models. The standard deviations of adversarial results are considerably larger than the reference models or visual blind models. Perturbations in the visual field make the predictions unstable.
These observations represent a positive sign, showing that RSVQA models extract some relevant information from the visual input and depend on it to predict an answer. The differences being stronger for fine-tuned models could indicate again that the process of fine-tuning might help in aligning textual embedding with visual embedding for better interactions. 
Nevertheless, the importance of language biases in RSVQA remains large and models can leverage and focus solely on language for a considerable proportion of samples. The issue is further illustrated with two examples in Figure~\ref{fig:ManualAdTest}, where two images are manipulated by removing or adding a large forest patch, and sent through the models along with a question about the presence of forest. Forest being a common feature in the region covered by the dataset (the Netherlands), a positive answer is more commonly associated with it. This is verified in Figure~\ref{fig:ManualAdTest}, as the answer remains the same with or without manipulation, regardless of the version of the model used. 

\subsection{Dataset analysis of RSVQA Low resolution}
In Section~\ref{subsec:blindAd_discussion}, we observed the effects of language bias. A clear understanding of the dataset is necessary to identify first-hand the imbalanced distribution of question-answer pairs. 

\begin{figure}[!t]
    \centering
    \includegraphics[width = 1\textwidth]{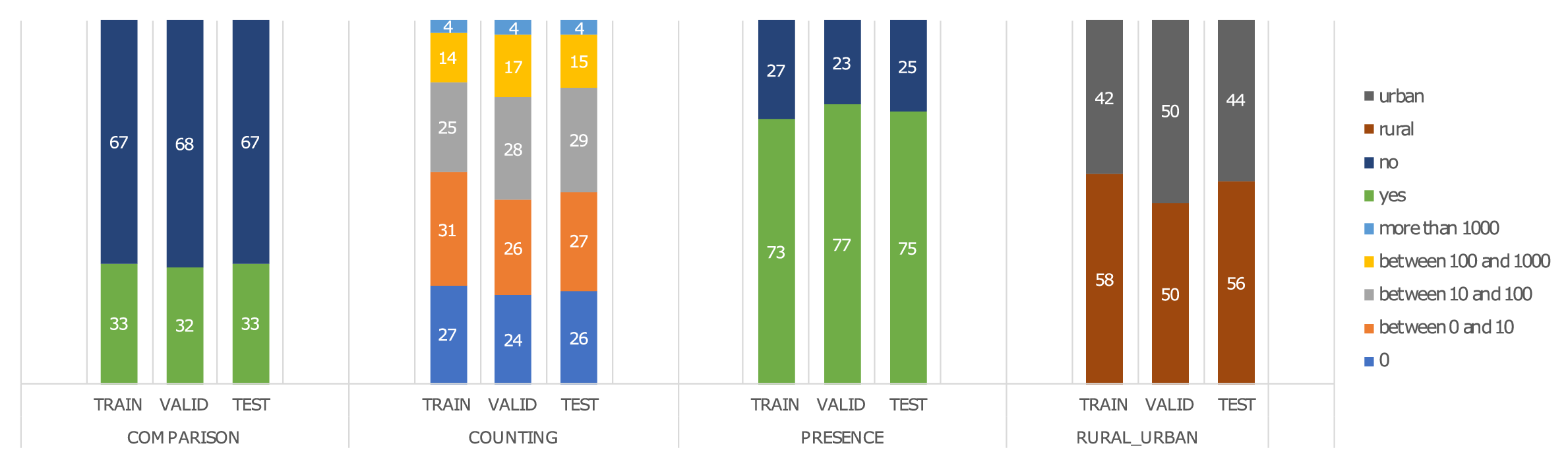}
    \caption{Distribution of answers for each question type and by dataset split (set of data for training, validation and testing). The percentage of each answer in the split is displayed. For each question type, the distribution is similar through the dataset splits.}
    \label{fig:data-distribution}
\end{figure}

The distribution of answers by question type and across the dataset splits is presented in Figure~\ref{fig:data-distribution}. ``Presence'' and ``Comparison'' questions are answered by yes/no. Numerical answers are re-framed as categorical ranges for ``Counting'' questions. ``Rural/urban'' questions, identifying the general category of the image, account for only 1\% of the dataset. This figure illustrates the language prior, the most common answer by question type, and is especially strong for ``Presence'' (73-75\% of yes) and ``Comparison'' (67-68\% of no) questions. This is coherent with results of the visual blind models discussed in Section~\ref{subsec:blindAd_discussion}.

As described in Section~\ref{sec:DA}, the dataset analysis is conducted in two steps. 
Applying the accuracy metrics detailed in Section~\ref{sec:DA} gives the results displayed in Table ~\ref{tab:high-level data analysis} and Table ~\ref{tab:low-level data analysis}. While the former illustrates the high-level analysis of a model predicting the most common answer or prior at the question-type level ($\mathbf{Prior}_{qType}$), a more in-depth analysis is performed in the latter, at the objects and patterns level. 

\begin{table}[!b]
    \caption{Performances on RSVQA LR test set when predicting the most common ($\mathbf{Prior\_Acc}$) or uniform ($\mathbf{Uni\_Acc}$) answers. "All" refers to the entire testing set. The average accuracy of the best model in this study, RSVQA with fine-tuned distilBERT, is displayed under $\mathbf{RSVQA\_Acc}$ while the proposed improvement/relative metrics are calculated in column $\mathbf{IO\_Prior}$ and $\mathbf{IO\_AdTest}$}
    \label{tab:high-level data analysis}
    \centering
    \resizebox{1\linewidth}{!}{%
    \begin{tabular}{|c|c|c|c|c|c|c|c|c|}
     \hline
     Question Type & $N_s$ & $|\mathcal{A}_q|$ & Most common answer & $\mathbf{Prior\_Acc}$ (\%) & $\mathbf{Uni\_Acc}$ (\%) & $\mathbf{RSVQA\_Acc} (\%)$ & $\mathbf{IO\_Prior} (\%)$ & $\mathbf{IO\_AdTest} (\%)$\\
     \hline
     All & 10'004 & 9 & yes & 35.47 & 11.11 &  & & \\
     \hline \hline
     Comparison & 4'002 & 2 & no & 66.74 & 50.00 & 90.76 & 72.22 & 54.84\\
     Counting & 2'947 & 5 & between 0 and 10 & 27.38 & 20.00 & 70.59 & 59.87 & 51.89\\
     Presence & 2'955 & 2 & yes & 75.03 & 50.00 & 91.04 & 64.12 & 66.79\\
     Rural/urban & 100 & 2 & rural & 56.00 & 50.00 & 90.00 & 77.27 & 77.27\\
     \hline
     Average & & & & 56.29 & 42.5 & 85.60 & 67.06 & 62.24\\
     Overall & & & & \textbf{57.49} & \textbf{41.04} & \textbf{84.90} & \textbf{64.48} &\textbf{56.36}\\
     \hline
    \end{tabular}}
\end{table}

\begin{table}[!b]
    \caption{The in-depth analysis of the RSVQA LR dataset is performed on objects and relations ($s = objects\&patterns$). The identification of the subset most common or uniformly distributed answers is done for different grouping strategy: \textit{pattern, object, object with attribute}, for example ``Is there a [...] ?'', ``water area'' and ``small water area'' respectively.}
    \label{tab:low-level data analysis}
    \centering
    \resizebox{0.5\linewidth}{!}{%
    \begin{tabular}{|c|c|c|c|}
     \hline
     Question Type & Group & $\mathbf{Prior\_Acc}$ (\%) & $\mathbf{Uni\_Acc}$ (\%) \\
     \hline
     \multirow{3}{*}{Comparison}& \textbf{pattern} & \textbf{67.29}  & \multirow{3}{*}{50.00} \\
     & object & 66.17 & \\
     & object (w/ attr) & 65.42  & \\
     \hline
     \multirow{3}{*}{Counting} & pattern & 27.38 & \multirow{3}{*}{20.00} \\
     & object & 44.59 &  \\
     & \textbf{object (w/ attr)} & \textbf{59.55} &  \\
     \hline
     \multirow{3}{*}{Presence} & pattern & 75.03 & \multirow{3}{*}{50.00} \\
     & object & 78.65 &  \\
     & \textbf{object (w/ attr)} & \textbf{89.04} & \\
     \hline
     Rural/urban & n/a & 56.00 & 50.00 \\
     \hline
     \multirow{3}{*}{Overall} & pattern & 57.71 & \multirow{3}{*}{41.04} \\
     & object & 63.39 &  \\
     & \textbf{object (w/ attr)} & \textbf{70.57} & \\
     \hline
    \end{tabular}}
\end{table}

The dataset analysis helps to gain a better understanding of its language bias. While Figure~\ref{fig:data-distribution} visually shows the imbalance for ``Comparison'' and ``Presence'' categories, Table~\ref{tab:high-level data analysis} translates it into $\mathbf{Prior\_Acc}_{qType}$, the accuracy of a model selecting the most common answer for each question type, and compares it to $\mathbf{Uni\_Acc}_{qType}$ assuming a uniform answer distribution, and $\mathbf{RSVQA\_Acc}$ our best model performances. Logically, we see that $\mathbf{Prior\_Acc}_{qType}$ correspond to the proportion of the most common answer per question type for the test set in Figure~\ref{fig:data-distribution}.
The new evaluation metrics, $\mathbf{IO\_Prior}$ as improvement from a dataset lower-bound and $\mathbf{IO\_AdTest}$ as the performance relative to improper visual information, gives a lower but more realistic idea of the capabilities of the baseline architecture with respect to the task and considering the bias data it was trained on. It also allows to highlight the considerable progress still open to new methodological development.

For a more detailed analysis, $\mathbf{Prior\_Acc}_{objects\&patterns}$ can be calculated, assuming a model identifying objects and patterns in the question. The dataset analysis at objects\&patterns level (Table~\ref{tab:low-level data analysis}) gives a higher accuracy metric, still without using any modern language model mechanisms. 
In particular, the $\mathbf{Prior\_Acc}$ of ``Counting'' and ``Presence'' questions with subsets defined by objects type with attributes are very close to the results of the visual blind models. Thus, the in-depth dataset analysis for these question types captures and illustrates well the language bias. For ``Comparison'' questions, the obtained $\mathbf{Prior\_Acc}$ are still well below the performance achieved with fine-tuned visual blind models. This indicates the additional complexity in this question type where more elements interacts: two objects, with or without attributes, with or without relations, that is not fully apprehended with the analysis performed. 
Globally, while this analysis shows in more details the imbalance, it is less replicable as the strategy to dissect a question can be slightly subjective.
While $\mathbf{Prior}_{objects\&patterns}$ is partially able to deconstruct the question construction scheme, it is still several points behind the deep learning architectures. However, it visually illustrates the challenge of language biases in a RSVQA dataset. When framing RSVQA as a classification task and evaluating the model with accuracy, it is important to raise awareness about the range of values this metric can really take and what is the minimal performance one should expect.

\subsection{Extension to other RSVQA datasets}
\label{subsec:DAresults_extension}

\begin{table}[!b]
    \caption{Dataset analysis (at the level of question type) of additional RSVQA datasets}
    \label{tab:Extension_datasetAnalysis}
    \centering
    \resizebox{0.95\linewidth}{!}{%
    \begin{tabular}{|c|c|c|c|c|c|c|}
    \hline
    Dataset & Question Type & $N_q$ & $|\mathcal{A}_q|$ & Most common answer & $\mathbf{Prior\_Acc}$ (\%) & $\mathbf{Uni\_Acc}$ (\%)\\
    \hline \hline
    
    \multirow{7}{*}{\shortstack{\textbf{RSVQA HR}\\ test 1}} & All & 222'684 & 55 & no & 37.81 & 1.82 \\
    \hhline {~------}
    & Area & 33'067 & 5 & 0m2 & 62.40 & 20.00 \\
    & Comparison & 72'923 & 2 & no & 66.62 & 50.00 \\
    & Counting & 58'149 & 48 & 0 & 60.64 & 2.08 \\
    & Presence & 58'545 & 2 & no & 60.82 & 50.00 \\
    \hhline {~------}
    & Average &  & & & \textbf{62.62} & \textbf{30.52} \\
    & Overall &  & & & \textbf{62.91} & \textbf{33.01} \\
    \hhline {-------}
    \multirow{7}{*}{\shortstack{\textbf{RSVQA HR}\\ test 2}} & All & 105'647 & 55 & no & 36.99 & 1.82 \\
    \hhline {~------}
    & Area & 15'535 & 5 & 0m2 & 57.42 & 20.00 \\
    & Comparison & 35'019 & 2 & no & 66.57 & 50.00 \\
    & Counting & 27'558 & 48 & 0 & 56.74 & 2.08 \\
    & Presence & 27'535 & 2 & no & 57.25 & 50.00 \\
    \hhline {~------}
    & Average &  & & & \textbf{59.50} & \textbf{30.52} \\
    & Overall &  & & & \textbf{60.23} & \textbf{33.09} \\
    \hline \hline
    
    \multirow{5}{*}{\shortstack{\textbf{RSVQAxBEN}\\ test}} & All & 2'953'125 & 26'875 & no & 51.96 & 0.004 \\
    \hhline {~------}
    & Land cover & 529'413 & 26'873 & None & 13.31 & 0.004 \\
    & Yes/No & 2'423'712 & 2 & no & 63.32 & 50.00 \\
    \hhline {~------}
    & Average & & &  & \textbf{38.31} & \textbf{25.00} \\
    & Overall & & &  & \textbf{54.35} & \textbf{41.04} \\
    \hline \hline
    
    \multirow{7}{*}{\textbf{RSIVQA}} & All & 111'693 & 579 & no & 29.03 & 0.17 \\
    \hhline {~------}
    & Counting & 32'331 & 472 & 1 & 48.01 & 0.21 \\
    & Location & 559 & 54 & center & 26.83 & 1.85 \\
    & Presence (Yes/No) & 64'610 & 2 & no & 50.19 & 50.00 \\
    & Scene & 14'193 & 52 & playfields & 7.26 & 1.92 \\
    \hhline {~------}
    & Average &  & & & \textbf{33.08} & \textbf{13.50} \\  
    & Overall &  & & & \textbf{43.99} & \textbf{29.24} \\  
    \hline \hline
    
    \multirow{9}{*}{\textbf{FloodNet}} & All & 4511 & 41 & non flooded & 43.23 & 2.44 \\
    \hhline {~------}
    & Simple Counting & 636 & 35 & 3 & 17.14 & 2.86 \\
    & Complex Counting & 693 & 35 & 3 & 17.89 & 2.86 \\
    & Condition Recognition - Yes/No & 867 & 2 & yes & 50.17 & 50.00 \\
    & Condition Recognition - Road & 867 & 3 & non flooded & 81.08 & 33.33 \\
    & Condition Recognition - Scene & 1448 & 2 & non flooded & 86.12 & 50.00 \\
    \hhline {~------}
    & Average & & & & \textbf{50.48} & \textbf{27.81} \\
    & Overall & & & & \textbf{58.03} & \textbf{32.91} \\
    \hhline {~------}
    & Question-level, overall & & & & \textbf{64.98} & \textbf{33.76}\\
    \hline
    
    \end{tabular}}
\end{table}

Extending this analysis to other RSVQA datasets (Table~\ref{tab:Extension_datasetAnalysis}) shows that a similar pattern can be observed across them. However, it is more pronounced in some cases than others, as we detail in this section. 
In RSVQA High resolution, results show an almost two-fold difference between the $\mathbf{Uni\_Acc}$ and $\mathbf{Prior\_Acc}$ scores. Therefore, the RSVQA High resolution dataset tends to be more unbalanced than the RSVQA Low resolution dataset. However, the second test set, which is located in a geographical area unseen in training, is slightly less proned to the language bias.
In FloodNet, as the number of unique questions is limited, $\mathbf{Prior\_Acc}$ and $\mathbf{Uni\_Acc}$ scores can be calculated as far as the level of single questions ($s=question$). The most common answer is searched for each question and assigned as prediction for all samples with this question in $\mathbf{Prior\_Acc}$. This calculation corresponds to the entry ``Question-level, overall'' in Table~\ref{tab:Extension_datasetAnalysis}, and is detailed in Appendix (Table~\ref{tab:FloodNet_low_level_analysis}).
Interestingly, the two datasets that are generated from existing resources (i.e. remote sensing dataset for other tasks), RSVQAxBEN and RSIVQA, have lower overall $\mathbf{Prior\_Acc}$. 
In RSIVQA, we imagine the existing datasets used (UCM~\citep{yang_bag--visual-words_2010}, Sydney~\citep{zhang_saliency-guided_2015}, AID~\citep{xia_aid_2017}, HRRSD~\citep{zhang_hierarchical_2019} and DOTA~\citep{xia_dota_2018}) already faced the question of balancing their samples. Being generated from them, RSIVQA probably inherits this property. Nevertheless, the lack of defined train/validation/test split cannot ensure the distribution of questions-answers through the sets, and remains a limitation for methods comparison.
RSVQAxBEN has the lowest difference between $\mathbf{Prior\_Acc}$ and $\mathbf{Uni\_Acc}$. While only two possible answers for Yes/No questions lead to a relatively high $\mathbf{Uni\_Acc}$, it is also possible that the domain gap between train/validation/test sets (being defined on latitude, rather than randomly) helps in mitigating language biases by differentiating the distribution of land covers across sets.

This analysis of existing datasets allows to identify sources of language biases that are specific to the field of remote sensing. Geographical sparsity is illustrated in particular by the frequency of answers such as ``0'', ``0m2'', ``no'', ``none''. As the absence of some elements is more frequent than their presence when observing the planet, models can naturally learn to directly predict negative or low number answers, without assessing the visual content. In parallel, existing datasets were developed with automatic procedures and a fix set of templates. This lack of vocabulary diversity is also a source of language bias. The case of FloodNet illustrates it well as it consists of only 15 different questions, and when considering each question individually with their most common answer, the obtained $\mathbf{Prior\_Acc}$ accuracy reaches about 65\%, regardless of the images. 


\section{Global discussion and Conclusion}
\label{sec:discussion_conclusion}

With this work, we wish to raise awareness around the challenges of language biases in RSVQA and propose relative, simple yet transparent, evaluation metrics to take it into account. Additionally, we encourage the community to push for research towards methods that enforce the grounding of the answer in the image, and more interpretable methods, to understand what triggered the model to produce a certain answer. Besides methodological progress, such future researches will need critical advances in the development of new types of datasets, carrying information with higher semantics, exceeding what can be achieved with the current image-question-answer triplets (for example scene graph as proposed by GQA~\citep{hudson_gqa_2019}). 

We analysed the language component, especially the language bias, in both models and datasets for remote sensing visual question answering. 
The language bias analysis is performed on a set of models with the same visual feature extractor but varying language models, with and without fine-tuning. In all cases (i.e. RNN or Transfomer), fine-tuning the model weights proved beneficial to achieve maximal performance as expected. While the simpler, traditional RNN performed already well with frozen weights, the distilBERT Transformer was slightly but almost consistently better when properly fine-tuned. While fine-tuning language models appeared as key, \citet{tsimpoukelli_multimodal_2021} raised some concerns by showing that keeping the language model frozen should help in retaining its few-shot learning abilities. Assessing the generalization capacities of models is left to future research as framing RSVQA as a classification problem represents a restriction for testing other datasets. Indeed, the answer classes are hard-encoded and defined during training, which limits the evaluation for new classes unseen during training. However, our results tend to indicate that the fine-tuning process helped to adapt the question representations to the task, leading to more efficient interactions with the image representations. Indeed, models with fine-tuned language extractors were more affected by the adversarial testing strategy in comparison to their frozen counter-part (both in terms of absolute values and variability). 

The training of visual blind models, as well as the dataset analysis, highlighted the large extent of the problem of language biases in RSVQA. Comparing with the same type of assessment in classical VQA, where language bias is well-recognized, we observe the issue is more severe in RSVQA, where visual blind models (also called ``Language-only'' in~\citet{goyal_making_2017}, ``d-LSTM Q'' in~\citet{agrawal_dont_2018} and ``LSTM'' in~\citet{kervadec_roses_2021}) are considerably higher in accuracy and closer to the corresponding reference model with access to both images and text. For instance, the difference between the visual blind ``d-LSTM Q'' model and the MCB model in~\citet{agrawal_dont_2018} is much more important on VQAv2 (16.7\% diffence in overall accuracy). Although the adversarial testing showed that models do extract visual information useful to predict the correct answer from this modality, there is an urgent need for the remote sensing community to start taking this challenge into account.

Any method of deep learning requires data to train, usually in large quantities. The characteristics of datasets can influence considerably results, as motivated in the data-centric movement in AI ~\citep{jarrahi_principles_2022}. 
We identified the importance of language biases in existing remote sensing VQA datasets, in particular how they relate to specific remote sensing characteristics, while realizing that bias can be a natural, inherent aspect of real-world data. We suggested reporting transparent evaluation metrics with respect to the potential accuracy gap to be filled. Considering only the data, the ``accuracy of the most common answer'' metric, $\mathbf{Prior\_Acc}$, should be clearly reported alongside new RSVQA datasets, and can be used in new method developments to report a dataset-based relative evaluation, alongside accuracy. In parallel, we suggest a second evaluation metric considering both dataset and method, the relative accuracy, using the adversarial testing results as a lower-bound performance. This metric has the advantage of considering the type of method (e.g. classification), and with limited extra computation cost. 

On the perspective of developing new datasets, the challenge of language biases should be of primary importance. The case of the ``Rural / Urban '' question type raised the issue of diversity, as this question shows the minimal effect of language bias, but does not reflect the motivation behind a VQA system, i.e. tackling a diversity of requests with a single system. 
The variety of answers is key as well. While ``yes/no'' questions were understandably a good first step in RSVQA, question types and samples with a wider range of possible answers and with a more balanced distribution are necessary. 
Language richness is essential to consider in dataset construction and challenges the concept of templates to create questions, which is used in most current RSVQA datasets. Indeed, it can be argued that modern language models can easily learn template-generated text. Moreover, the challenge remains to develop seemingly human questions, that would make sense for a human. 
Additionally, a human baseline might be an interesting and useful experiment to compare model performances with. How such baseline should be designed, both in terms of cohort of participants, themes and evaluation of the accuracy of human responses should be investigated in future research.


\section{Acknowledgement}
This work is supported by the European Space Agency through the Discovery and Preparation Program, and is part of the project
“An AI assistant to interact with remote sensing images” (ESA Contract no. 4000133903/21/NL/MH/hm) led in partnership between the ECEO lab (EPFL) and the $\Phi$-lab (ESA). \\
The authors would like to thank Ciprian Tomoiagă and Anthony Guinchard for the interesting exchanges and discussions on deep learning applied to the understanding of satellite imagery. 


\bibliographystyle{elsarticle-harv}
\biboptions{authoryear}
\bibliography{main.bib}

\begin{thebibliography}{65}
\expandafter\ifx\csname natexlab\endcsname\relax\def\natexlab#1{#1}\fi
\providecommand{\url}[1]{\texttt{#1}}
\providecommand{\href}[2]{#2}
\providecommand{\path}[1]{#1}
\providecommand{\DOIprefix}{doi:}
\providecommand{\ArXivprefix}{arXiv:}
\providecommand{\URLprefix}{URL: }
\providecommand{\Pubmedprefix}{pmid:}
\providecommand{\doi}[1]{\href{http://dx.doi.org/#1}{\path{#1}}}
\providecommand{\Pubmed}[1]{\href{pmid:#1}{\path{#1}}}
\providecommand{\bibinfo}[2]{#2}
\ifx\xfnm\relax \def\xfnm[#1]{\unskip,\space#1}\fi
\bibitem[{Agrawal et~al.(2018)Agrawal, Batra, Parikh and Kembhavi}]{agrawal_dont_2018}
\bibinfo{author}{Agrawal, A.}, \bibinfo{author}{Batra, D.}, \bibinfo{author}{Parikh, D.}, \bibinfo{author}{Kembhavi, A.}, \bibinfo{year}{2018}.
\newblock \bibinfo{title}{Don't {Just} {Assume}; {Look} and {Answer}: {Overcoming} {Priors} for {Visual} {Question} {Answering}}, in: \bibinfo{booktitle}{2018 {IEEE}/{CVF} {Conference} on {Computer} {Vision} and {Pattern} {Recognition}}, \bibinfo{publisher}{IEEE}, \bibinfo{address}{Salt Lake City, UT}. pp. \bibinfo{pages}{4971--4980}.
\newblock \URLprefix \url{https://ieeexplore.ieee.org/document/8578620/}, \DOIprefix\doi{10.1109/CVPR.2018.00522}.
\bibitem[{Anderson et~al.(2018)Anderson, He, Buehler, Teney, Johnson, Gould and Zhang}]{anderson_bottom-up_2018}
\bibinfo{author}{Anderson, P.}, \bibinfo{author}{He, X.}, \bibinfo{author}{Buehler, C.}, \bibinfo{author}{Teney, D.}, \bibinfo{author}{Johnson, M.}, \bibinfo{author}{Gould, S.}, \bibinfo{author}{Zhang, L.}, \bibinfo{year}{2018}.
\newblock \bibinfo{title}{Bottom-{Up} and {Top}-{Down} {Attention} for {Image} {Captioning} and {Visual} {Question} {Answering}}, in: \bibinfo{booktitle}{2018 {IEEE}/{CVF} {Conference} on {Computer} {Vision} and {Pattern} {Recognition}}, \bibinfo{publisher}{IEEE}, \bibinfo{address}{Salt Lake City, UT}. pp. \bibinfo{pages}{6077--6086}.
\newblock \URLprefix \url{https://ieeexplore.ieee.org/document/8578734/}, \DOIprefix\doi{10.1109/CVPR.2018.00636}.
\bibitem[{Andreas et~al.(2016)Andreas, Rohrbach, Darrell and Klein}]{andreas_neural_2016}
\bibinfo{author}{Andreas, J.}, \bibinfo{author}{Rohrbach, M.}, \bibinfo{author}{Darrell, T.}, \bibinfo{author}{Klein, D.}, \bibinfo{year}{2016}.
\newblock \bibinfo{title}{Neural {Module} {Networks}}, in: \bibinfo{booktitle}{2016 {IEEE} {Conference} on {Computer} {Vision} and {Pattern} {Recognition} ({CVPR})}, \bibinfo{publisher}{IEEE}, \bibinfo{address}{Las Vegas, NV, USA}. pp. \bibinfo{pages}{39--48}.
\newblock \URLprefix \url{http://ieeexplore.ieee.org/document/7780381/}, \DOIprefix\doi{10.1109/CVPR.2016.12}.
\bibitem[{Antol et~al.(2015)Antol, Agrawal, Lu, Mitchell, Batra, Zitnick and Parikh}]{antol_vqa_2015}
\bibinfo{author}{Antol, S.}, \bibinfo{author}{Agrawal, A.}, \bibinfo{author}{Lu, J.}, \bibinfo{author}{Mitchell, M.}, \bibinfo{author}{Batra, D.}, \bibinfo{author}{Zitnick, C.L.}, \bibinfo{author}{Parikh, D.}, \bibinfo{year}{2015}.
\newblock \bibinfo{title}{{VQA}: {Visual} {Question} {Answering}}, pp. \bibinfo{pages}{2425--2433}.
\newblock \URLprefix \url{https://openaccess.thecvf.com/content_iccv_2015/html/Antol_VQA_Visual_Question_ICCV_2015_paper.html}.
\bibitem[{Bazi et~al.(2022)Bazi, Rahhal, Mekhalfi, Zuair and Melgani}]{bazi_bi-modal_2022}
\bibinfo{author}{Bazi, Y.}, \bibinfo{author}{Rahhal, M.M.A.}, \bibinfo{author}{Mekhalfi, M.L.}, \bibinfo{author}{Zuair, M.A.A.}, \bibinfo{author}{Melgani, F.}, \bibinfo{year}{2022}.
\newblock \bibinfo{title}{Bi-{Modal} {Transformer}-{Based} {Approach} for {Visual} {Question} {Answering} in {Remote} {Sensing} {Imagery}}.
\newblock \bibinfo{journal}{IEEE Transactions on Geoscience and Remote Sensing} \bibinfo{volume}{60}, \bibinfo{pages}{1--11}.
\newblock \URLprefix \url{https://ieeexplore.ieee.org/document/9832935/}, \DOIprefix\doi{10.1109/TGRS.2022.3192460}.
\bibitem[{Ben-younes et~al.(2017)Ben-younes, Cadene, Cord and Thome}]{ben-younes_mutan_2017}
\bibinfo{author}{Ben-younes, H.}, \bibinfo{author}{Cadene, R.}, \bibinfo{author}{Cord, M.}, \bibinfo{author}{Thome, N.}, \bibinfo{year}{2017}.
\newblock \bibinfo{title}{{MUTAN}: {Multimodal} {Tucker} {Fusion} for {Visual} {Question} {Answering}}, in: \bibinfo{booktitle}{2017 {IEEE} {International} {Conference} on {Computer} {Vision} ({ICCV})}, \bibinfo{publisher}{IEEE}, \bibinfo{address}{Venice}. pp. \bibinfo{pages}{2631--2639}.
\newblock \URLprefix \url{http://ieeexplore.ieee.org/document/8237547/}, \DOIprefix\doi{10.1109/ICCV.2017.285}.
\bibitem[{Bhargava and Ng(2022)}]{bhargava_commonsense_2022}
\bibinfo{author}{Bhargava, P.}, \bibinfo{author}{Ng, V.}, \bibinfo{year}{2022}.
\newblock \bibinfo{title}{Commonsense {Knowledge} {Reasoning} and {Generation} with {Pre}-trained {Language} {Models}: {A} {Survey}}.
\newblock \bibinfo{journal}{Proceedings of the AAAI Conference on Artificial Intelligence} \bibinfo{volume}{36}, \bibinfo{pages}{12317--12325}.
\newblock \URLprefix \url{https://ojs.aaai.org/index.php/AAAI/article/view/21496}, \DOIprefix\doi{10.1609/aaai.v36i11.21496}.
\bibitem[{Brown et~al.(2020)Brown, Mann, Ryder, Subbiah, Kaplan, Dhariwal, Neelakantan, Shyam, Sastry, Askell, Agarwal, Herbert-Voss, Krueger, Henighan, Child, Ramesh, Ziegler, Wu, Winter, Hesse, Chen, Sigler, Litwin, Gray, Chess, Clark, Berner, McCandlish, Radford, Sutskever and Amodei}]{brown_language_2020}
\bibinfo{author}{Brown, T.}, \bibinfo{author}{Mann, B.}, \bibinfo{author}{Ryder, N.}, \bibinfo{author}{Subbiah, M.}, \bibinfo{author}{Kaplan, J.D.}, \bibinfo{author}{Dhariwal, P.}, \bibinfo{author}{Neelakantan, A.}, \bibinfo{author}{Shyam, P.}, \bibinfo{author}{Sastry, G.}, \bibinfo{author}{Askell, A.}, \bibinfo{author}{Agarwal, S.}, \bibinfo{author}{Herbert-Voss, A.}, \bibinfo{author}{Krueger, G.}, \bibinfo{author}{Henighan, T.}, \bibinfo{author}{Child, R.}, \bibinfo{author}{Ramesh, A.}, \bibinfo{author}{Ziegler, D.}, \bibinfo{author}{Wu, J.}, \bibinfo{author}{Winter, C.}, \bibinfo{author}{Hesse, C.}, \bibinfo{author}{Chen, M.}, \bibinfo{author}{Sigler, E.}, \bibinfo{author}{Litwin, M.}, \bibinfo{author}{Gray, S.}, \bibinfo{author}{Chess, B.}, \bibinfo{author}{Clark, J.}, \bibinfo{author}{Berner, C.}, \bibinfo{author}{McCandlish, S.}, \bibinfo{author}{Radford, A.}, \bibinfo{author}{Sutskever, I.}, \bibinfo{author}{Amodei, D.}, \bibinfo{year}{2020}.
\newblock \bibinfo{title}{Language {Models} are {Few}-{Shot} {Learners}}, in: \bibinfo{editor}{Larochelle, H.}, \bibinfo{editor}{Ranzato, M.}, \bibinfo{editor}{Hadsell, R.}, \bibinfo{editor}{Balcan, M.F.}, \bibinfo{editor}{Lin, H.} (Eds.), \bibinfo{booktitle}{Advances in {Neural} {Information} {Processing} {Systems}}, \bibinfo{publisher}{Curran Associates, Inc.}. pp. \bibinfo{pages}{1877--1901}.
\newblock \URLprefix \url{https://proceedings.neurips.cc/paper/2020/file/1457c0d6bfcb4967418bfb8ac142f64a-Paper.pdf}.
\bibitem[{Camps‐Valls et~al.(2021)Camps‐Valls, Tuia, Zhu and Reichstein}]{campsvalls_deep_2021}
\bibinfo{editor}{Camps‐Valls, G.}, \bibinfo{editor}{Tuia, D.}, \bibinfo{editor}{Zhu, X.X.}, \bibinfo{editor}{Reichstein, M.} (Eds.), \bibinfo{year}{2021}.
\newblock \bibinfo{title}{Deep {Learning} for the {Earth} {Sciences}: {A} {Comprehensive} {Approach} to {Remote} {Sensing}, {Climate} {Science}, and {Geosciences}}.
\newblock \bibinfo{edition}{1} ed., \bibinfo{publisher}{Wiley}.
\newblock \URLprefix \url{https://onlinelibrary.wiley.com/doi/book/10.1002/9781119646181}, \DOIprefix\doi{10.1002/9781119646181}.
\bibitem[{Cao et~al.(2022)Cao, Li, Li, Nie and Zhang}]{cao_image-text_2022}
\bibinfo{author}{Cao, M.}, \bibinfo{author}{Li, S.}, \bibinfo{author}{Li, J.}, \bibinfo{author}{Nie, L.}, \bibinfo{author}{Zhang, M.}, \bibinfo{year}{2022}.
\newblock \bibinfo{title}{Image-text {Retrieval}: {A} {Survey} on {Recent} {Research} and {Development}}, in: \bibinfo{booktitle}{Proceedings of the {Thirty}-{First} {International} {Joint} {Conference} on {Artificial} {Intelligence}}, \bibinfo{publisher}{International Joint Conferences on Artificial Intelligence Organization}, \bibinfo{address}{Vienna, Austria}. pp. \bibinfo{pages}{5410--5417}.
\newblock \URLprefix \url{https://www.ijcai.org/proceedings/2022/759}, \DOIprefix\doi{10.24963/ijcai.2022/759}.
\bibitem[{Chappuis et~al.(2021)Chappuis, Lobry, Kellenberger, Le~Saux and Tuia}]{chappuis_how_2021}
\bibinfo{author}{Chappuis, C.}, \bibinfo{author}{Lobry, S.}, \bibinfo{author}{Kellenberger, B.}, \bibinfo{author}{Le~Saux, B.}, \bibinfo{author}{Tuia, D.}, \bibinfo{year}{2021}.
\newblock \bibinfo{title}{How to find a good image-text embedding for remote sensing visual question answering?}, in: \bibinfo{booktitle}{{ECML}-{PKDD} 2021 ({MACLEAN} workshop)}.
\newblock \URLprefix \url{http://arxiv.org/abs/2109.11848}. \bibinfo{note}{arXiv: 2109.11848}.
\bibitem[{Chappuis et~al.(2022a)Chappuis, Mendez, Walt, Lobry, Le~Saux and Tuia}]{chappuis_language_2022}
\bibinfo{author}{Chappuis, C.}, \bibinfo{author}{Mendez, V.}, \bibinfo{author}{Walt, E.}, \bibinfo{author}{Lobry, S.}, \bibinfo{author}{Le~Saux, B.}, \bibinfo{author}{Tuia, D.}, \bibinfo{year}{2022}a.
\newblock \bibinfo{title}{Language {Transformers} for {Remote} {Sensing} {Visual} {Question} {Answering}}, in: \bibinfo{booktitle}{{IGARSS} 2022 - 2022 {IEEE} {International} {Geoscience} and {Remote} {Sensing} {Symposium}}, \bibinfo{publisher}{IEEE}, \bibinfo{address}{Kuala Lumpur, Malaysia}. pp. \bibinfo{pages}{4855--4858}.
\newblock \URLprefix \url{https://ieeexplore.ieee.org/document/9884036/}, \DOIprefix\doi{10.1109/IGARSS46834.2022.9884036}.
\bibitem[{Chappuis et~al.(2023)Chappuis, Sertic, Santacroce, Castillo~Navarro, Lobry, Le~Saux and Tuia}]{christelmveo}
\bibinfo{author}{Chappuis, C.}, \bibinfo{author}{Sertic, C.}, \bibinfo{author}{Santacroce, N.}, \bibinfo{author}{Castillo~Navarro, J.}, \bibinfo{author}{Lobry, S.}, \bibinfo{author}{Le~Saux, B.}, \bibinfo{author}{Tuia, D.}, \bibinfo{year}{2023}.
\newblock \bibinfo{title}{Multi-task prompt-rsvqa to explicitly count objects on aerial images}, in: \bibinfo{booktitle}{Workshop on Machine Vision for Earth Observation (MVEO) at the 34th British Machine Vision Conference (BMVC)}.
\bibitem[{Chappuis et~al.(2022b)Chappuis, Zermatten, Lobry, Le~Saux and Tuia}]{chappuis_prompt-rsvqa_2022}
\bibinfo{author}{Chappuis, C.}, \bibinfo{author}{Zermatten, V.}, \bibinfo{author}{Lobry, S.}, \bibinfo{author}{Le~Saux, B.}, \bibinfo{author}{Tuia, D.}, \bibinfo{year}{2022}b.
\newblock \bibinfo{title}{Prompt-{RSVQA}: {Prompting} {Visual} {Context} to a {Language} {Model} for {Remote} {Sensing} {Visual} {Question} {Answering}}, pp. \bibinfo{pages}{1372--1381}.
\newblock \URLprefix \url{https://openaccess.thecvf.com/content/CVPR2022W/EarthVision/html/Chappuis_Prompt-RSVQA_Prompting_Visual_Context_to_a_Language_Model_for_Remote_CVPRW_2022_paper.html}.
\bibitem[{Chen et~al.(2020)Chen, Li, Yu, El~Kholy, Ahmed, Gan, Cheng and Liu}]{chen_uniter_2020}
\bibinfo{author}{Chen, Y.C.}, \bibinfo{author}{Li, L.}, \bibinfo{author}{Yu, L.}, \bibinfo{author}{El~Kholy, A.}, \bibinfo{author}{Ahmed, F.}, \bibinfo{author}{Gan, Z.}, \bibinfo{author}{Cheng, Y.}, \bibinfo{author}{Liu, J.}, \bibinfo{year}{2020}.
\newblock \bibinfo{title}{{UNITER}: {UNiversal} {Image}-{TExt} {Representation} {Learning}}, in: \bibinfo{editor}{Vedaldi, A.}, \bibinfo{editor}{Bischof, H.}, \bibinfo{editor}{Brox, T.}, \bibinfo{editor}{Frahm, J.M.} (Eds.), \bibinfo{booktitle}{Computer {Vision} – {ECCV} 2020}, \bibinfo{publisher}{Springer International Publishing}, \bibinfo{address}{Cham}. pp. \bibinfo{pages}{104--120}.
\newblock \DOIprefix\doi{10.1007/978-3-030-58577-8_7}.
\bibitem[{Cho et~al.(2014)Cho, van Merrienboer, Gulcehre, Bahdanau, Bougares, Schwenk and Bengio}]{cho_learning_2014}
\bibinfo{author}{Cho, K.}, \bibinfo{author}{van Merrienboer, B.}, \bibinfo{author}{Gulcehre, C.}, \bibinfo{author}{Bahdanau, D.}, \bibinfo{author}{Bougares, F.}, \bibinfo{author}{Schwenk, H.}, \bibinfo{author}{Bengio, Y.}, \bibinfo{year}{2014}.
\newblock \bibinfo{title}{Learning {Phrase} {Representations} using {RNN} {Encoder}–{Decoder} for {Statistical} {Machine} {Translation}}, in: \bibinfo{booktitle}{Proceedings of the 2014 {Conference} on {Empirical} {Methods} in {Natural} {Language} {Processing} ({EMNLP})}, \bibinfo{publisher}{Association for Computational Linguistics}, \bibinfo{address}{Doha, Qatar}. pp. \bibinfo{pages}{1724--1734}.
\newblock \URLprefix \url{http://aclweb.org/anthology/D14-1179}, \DOIprefix\doi{10.3115/v1/D14-1179}.
\bibitem[{Deng et~al.(2009)Deng, Dong, Socher, Li, {Kai Li} and {Li Fei-Fei}}]{deng_imagenet_2009}
\bibinfo{author}{Deng, J.}, \bibinfo{author}{Dong, W.}, \bibinfo{author}{Socher, R.}, \bibinfo{author}{Li, L.J.}, \bibinfo{author}{{Kai Li}}, \bibinfo{author}{{Li Fei-Fei}}, \bibinfo{year}{2009}.
\newblock \bibinfo{title}{{ImageNet}: {A} large-scale hierarchical image database}, in: \bibinfo{booktitle}{2009 {IEEE} {Conference} on {Computer} {Vision} and {Pattern} {Recognition}}, \bibinfo{publisher}{IEEE}, \bibinfo{address}{Miami, FL}. pp. \bibinfo{pages}{248--255}.
\newblock \URLprefix \url{https://ieeexplore.ieee.org/document/5206848/}, \DOIprefix\doi{10.1109/CVPR.2009.5206848}.
\bibitem[{Devlin et~al.(2019)Devlin, Chang, Lee and Toutanova}]{devlin_bert_2019}
\bibinfo{author}{Devlin, J.}, \bibinfo{author}{Chang, M.W.}, \bibinfo{author}{Lee, K.}, \bibinfo{author}{Toutanova, K.}, \bibinfo{year}{2019}.
\newblock \bibinfo{title}{{BERT}: {Pre}-training of {Deep} {Bidirectional} {Transformers} for {Language} {Understanding}}, in: \bibinfo{booktitle}{Proceedings of the 2019 {Conference} of the {North} {American} {Chapter} of the {Association} for {Computational} {Linguistics}: {Human} {Language} {Technologies}, {Volume} 1 ({Long} and {Short} {Papers})}, \bibinfo{publisher}{Association for Computational Linguistics}, \bibinfo{address}{Minneapolis, Minnesota}. pp. \bibinfo{pages}{4171--4186}.
\newblock \URLprefix \url{http://aclweb.org/anthology/N19-1423}, \DOIprefix\doi{10.18653/v1/N19-1423}.
\bibitem[{Faure et~al.(2022)Faure, Lobry, Kurtz and Wendling}]{faure_embedding_2022}
\bibinfo{author}{Faure, M.}, \bibinfo{author}{Lobry, S.}, \bibinfo{author}{Kurtz, C.}, \bibinfo{author}{Wendling, L.}, \bibinfo{year}{2022}.
\newblock \bibinfo{title}{Embedding {Spatial} {Relations} in {Visual} {Question} {Answering} for {Remote} {Sensing}}, in: \bibinfo{booktitle}{2022 26th {International} {Conference} on {Pattern} {Recognition} ({ICPR})}, \bibinfo{publisher}{IEEE}, \bibinfo{address}{Montreal, QC, Canada}. pp. \bibinfo{pages}{310--316}.
\newblock \URLprefix \url{https://ieeexplore.ieee.org/document/9956401/}, \DOIprefix\doi{10.1109/ICPR56361.2022.9956401}.
\bibitem[{Felix et~al.(2021)Felix, Repasky, Hodge, Zolfaghari, Abbasnejad and Sherrah}]{felix_cross-modal_2021}
\bibinfo{author}{Felix, R.}, \bibinfo{author}{Repasky, B.}, \bibinfo{author}{Hodge, S.}, \bibinfo{author}{Zolfaghari, R.}, \bibinfo{author}{Abbasnejad, E.}, \bibinfo{author}{Sherrah, J.}, \bibinfo{year}{2021}.
\newblock \bibinfo{title}{Cross-{Modal} {Visual} {Question} {Answering} for {Remote} {Sensing} {Data}}, in: \bibinfo{booktitle}{2021 {Digital} {Image} {Computing}: {Techniques} and {Applications} ({DICTA})}, \bibinfo{publisher}{IEEE}, \bibinfo{address}{Gold Coast, Australia}. pp. \bibinfo{pages}{1--9}.
\newblock \URLprefix \url{https://ieeexplore.ieee.org/document/9647287/}, \DOIprefix\doi{10.1109/DICTA52665.2021.9647287}.
\bibitem[{Goyal et~al.(2017)Goyal, Khot, Summers-Stay, Batra and Parikh}]{goyal_making_2017}
\bibinfo{author}{Goyal, Y.}, \bibinfo{author}{Khot, T.}, \bibinfo{author}{Summers-Stay, D.}, \bibinfo{author}{Batra, D.}, \bibinfo{author}{Parikh, D.}, \bibinfo{year}{2017}.
\newblock \bibinfo{title}{Making the v in {VQA} {Matter}: {Elevating} the {Role} of {Image} {Understanding} in {Visual} {Question} {Answering}}, pp. \bibinfo{pages}{6904--6913}.
\newblock \URLprefix \url{https://openaccess.thecvf.com/content_cvpr_2017/html/Goyal_Making_the_v_CVPR_2017_paper.html}.
\bibitem[{Gupta et~al.(2022)Gupta, Li, Kortylewski, Zhang, Li and Yuille}]{gupta_swapmix_2022}
\bibinfo{author}{Gupta, V.}, \bibinfo{author}{Li, Z.}, \bibinfo{author}{Kortylewski, A.}, \bibinfo{author}{Zhang, C.}, \bibinfo{author}{Li, Y.}, \bibinfo{author}{Yuille, A.}, \bibinfo{year}{2022}.
\newblock \bibinfo{title}{{SwapMix}: {Diagnosing} and {Regularizing} the {Over}-{Reliance} on {Visual} {Context} in {Visual} {Question} {Answering}}, in: \bibinfo{booktitle}{2022 {IEEE}/{CVF} {Conference} on {Computer} {Vision} and {Pattern} {Recognition} ({CVPR})}, \bibinfo{publisher}{IEEE}, \bibinfo{address}{New Orleans, LA, USA}. pp. \bibinfo{pages}{5068--5078}.
\newblock \URLprefix \url{https://ieeexplore.ieee.org/document/9879464/}, \DOIprefix\doi{10.1109/CVPR52688.2022.00502}.
\bibitem[{He et~al.(2016)He, Zhang, Ren and Sun}]{he_deep_2016}
\bibinfo{author}{He, K.}, \bibinfo{author}{Zhang, X.}, \bibinfo{author}{Ren, S.}, \bibinfo{author}{Sun, J.}, \bibinfo{year}{2016}.
\newblock \bibinfo{title}{Deep {Residual} {Learning} for {Image} {Recognition}}, pp. \bibinfo{pages}{770--778}.
\newblock \URLprefix \url{https://openaccess.thecvf.com/content_cvpr_2016/html/He_Deep_Residual_Learning_CVPR_2016_paper.html}.
\bibitem[{Hinton et~al.(2015)Hinton, Vinyals and Dean}]{hinton_distilling_2015}
\bibinfo{author}{Hinton, G.}, \bibinfo{author}{Vinyals, O.}, \bibinfo{author}{Dean, J.}, \bibinfo{year}{2015}.
\newblock \bibinfo{title}{Distilling the {Knowledge} in a {Neural} {Network}}, in: \bibinfo{booktitle}{The proceedings of {Deep} {Learning} {Worshop}}, \bibinfo{publisher}{arXiv}.
\newblock \URLprefix \url{http://arxiv.org/abs/1503.02531}. \bibinfo{note}{arXiv:1503.02531 [cs, stat]}.
\bibitem[{Hudson and Manning(2019)}]{hudson_gqa_2019}
\bibinfo{author}{Hudson, D.A.}, \bibinfo{author}{Manning, C.D.}, \bibinfo{year}{2019}.
\newblock \bibinfo{title}{{GQA}: {A} {New} {Dataset} for {Real}-{World} {Visual} {Reasoning} and {Compositional} {Question} {Answering}}, pp. \bibinfo{pages}{6700--6709}.
\newblock \URLprefix \url{https://openaccess.thecvf.com/content_CVPR_2019/html/Hudson_GQA_A_New_Dataset_for_Real-World_Visual_Reasoning_and_Compositional_CVPR_2019_paper.html}.
\bibitem[{Jarrahi et~al.(2022)Jarrahi, Memariani and Guha}]{jarrahi_principles_2022}
\bibinfo{author}{Jarrahi, M.H.}, \bibinfo{author}{Memariani, A.}, \bibinfo{author}{Guha, S.}, \bibinfo{year}{2022}.
\newblock \bibinfo{title}{The {Principles} of {Data}-{Centric} {AI} ({DCAI})} \URLprefix \url{https://arxiv.org/abs/2211.14611}, \DOIprefix\doi{10.48550/ARXIV.2211.14611}. \bibinfo{note}{publisher: arXiv Version Number: 1}.
\bibitem[{Johnson et~al.(2017)Johnson, Hariharan, van~der Maaten, Fei-Fei, Zitnick and Girshick}]{johnson_clevr_2017}
\bibinfo{author}{Johnson, J.}, \bibinfo{author}{Hariharan, B.}, \bibinfo{author}{van~der Maaten, L.}, \bibinfo{author}{Fei-Fei, L.}, \bibinfo{author}{Zitnick, C.L.}, \bibinfo{author}{Girshick, R.}, \bibinfo{year}{2017}.
\newblock \bibinfo{title}{{CLEVR}: {A} {Diagnostic} {Dataset} for {Compositional} {Language} and {Elementary} {Visual} {Reasoning}}, in: \bibinfo{booktitle}{2017 {IEEE} {Conference} on {Computer} {Vision} and {Pattern} {Recognition} ({CVPR})}, \bibinfo{publisher}{IEEE}, \bibinfo{address}{Honolulu, HI}. pp. \bibinfo{pages}{1988--1997}.
\newblock \URLprefix \url{https://ieeexplore.ieee.org/document/8099698/}, \DOIprefix\doi{10.1109/CVPR.2017.215}.
\bibitem[{Kervadec et~al.(2021)Kervadec, Antipov, Baccouche and Wolf}]{kervadec_roses_2021}
\bibinfo{author}{Kervadec, C.}, \bibinfo{author}{Antipov, G.}, \bibinfo{author}{Baccouche, M.}, \bibinfo{author}{Wolf, C.}, \bibinfo{year}{2021}.
\newblock \bibinfo{title}{Roses are {Red}, {Violets} are {Blue}… {But} {Should} {VQA} expect {Them} {To}?}, in: \bibinfo{booktitle}{2021 {IEEE}/{CVF} {Conference} on {Computer} {Vision} and {Pattern} {Recognition} ({CVPR})}, \bibinfo{publisher}{IEEE}, \bibinfo{address}{Nashville, TN, USA}. pp. \bibinfo{pages}{2775--2784}.
\newblock \URLprefix \url{https://ieeexplore.ieee.org/document/9578217/}, \DOIprefix\doi{10.1109/CVPR46437.2021.00280}.
\bibitem[{Kingma and Ba(2015)}]{kingma_adam_2015}
\bibinfo{author}{Kingma, D.P.}, \bibinfo{author}{Ba, J.}, \bibinfo{year}{2015}.
\newblock \bibinfo{title}{Adam: {A} {Method} for {Stochastic} {Optimization}}, in: \bibinfo{booktitle}{International {Conference} on {Learning} {Representations} ({ICLR} 2015)}, \bibinfo{address}{San Diego, California}.
\newblock \URLprefix \url{http://arxiv.org/abs/1412.6980}. \bibinfo{note}{arXiv: 1412.6980}.
\bibitem[{Kiros et~al.(2015)Kiros, Zhu, Salakhutdinov, Zemel, Torralba, Urtasun and Fidler}]{kiros_skip-thought_2015}
\bibinfo{author}{Kiros, R.}, \bibinfo{author}{Zhu, Y.}, \bibinfo{author}{Salakhutdinov, R.}, \bibinfo{author}{Zemel, R.S.}, \bibinfo{author}{Torralba, A.}, \bibinfo{author}{Urtasun, R.}, \bibinfo{author}{Fidler, S.}, \bibinfo{year}{2015}.
\newblock \bibinfo{title}{Skip-{Thought} {Vectors}}, in: \bibinfo{booktitle}{{arXiv}:1506.06726 [cs]}.
\newblock \URLprefix \url{http://arxiv.org/abs/1506.06726}. \bibinfo{note}{arXiv: 1506.06726}.
\bibitem[{Krishna et~al.(2017)Krishna, Zhu, Groth, Johnson, Hata, Kravitz, Chen, Kalantidis, Li, Shamma, Bernstein and Fei-Fei}]{krishna_visual_2017}
\bibinfo{author}{Krishna, R.}, \bibinfo{author}{Zhu, Y.}, \bibinfo{author}{Groth, O.}, \bibinfo{author}{Johnson, J.}, \bibinfo{author}{Hata, K.}, \bibinfo{author}{Kravitz, J.}, \bibinfo{author}{Chen, S.}, \bibinfo{author}{Kalantidis, Y.}, \bibinfo{author}{Li, L.J.}, \bibinfo{author}{Shamma, D.A.}, \bibinfo{author}{Bernstein, M.S.}, \bibinfo{author}{Fei-Fei, L.}, \bibinfo{year}{2017}.
\newblock \bibinfo{title}{Visual {Genome}: {Connecting} {Language} and {Vision} {Using} {Crowdsourced} {Dense} {Image} {Annotations}}.
\newblock \bibinfo{journal}{International Journal of Computer Vision} \bibinfo{volume}{123}, \bibinfo{pages}{32--73}.
\newblock \URLprefix \url{http://link.springer.com/10.1007/s11263-016-0981-7}, \DOIprefix\doi{10.1007/s11263-016-0981-7}.
\bibitem[{Lachezar et~al.(2018)Lachezar, Lyubka, Vasil and Stuart}]{lachezar_challenges_2018}
\bibinfo{author}{Lachezar, F.}, \bibinfo{author}{Lyubka, P.}, \bibinfo{author}{Vasil, K.}, \bibinfo{author}{Stuart, F.}, \bibinfo{year}{2018}.
\newblock \bibinfo{title}{Challenges and {Solutions} for {Utilizing} {Earth} {Observations} in the "{Big} {Data}" era} \URLprefix \url{https://zenodo.org/record/2391936}, \DOIprefix\doi{10.5281/ZENODO.2391936}. \bibinfo{note}{publisher: Zenodo}.
\bibitem[{Lin et~al.(2014)Lin, Maire, Belongie, Hays, Perona, Ramanan, Dollár and Zitnick}]{fleet_microsoft_2014}
\bibinfo{author}{Lin, T.Y.}, \bibinfo{author}{Maire, M.}, \bibinfo{author}{Belongie, S.}, \bibinfo{author}{Hays, J.}, \bibinfo{author}{Perona, P.}, \bibinfo{author}{Ramanan, D.}, \bibinfo{author}{Dollár, P.}, \bibinfo{author}{Zitnick, C.L.}, \bibinfo{year}{2014}.
\newblock \bibinfo{title}{Microsoft {COCO}: {Common} {Objects} in {Context}}, in: \bibinfo{editor}{Fleet, D.}, \bibinfo{editor}{Pajdla, T.}, \bibinfo{editor}{Schiele, B.}, \bibinfo{editor}{Tuytelaars, T.} (Eds.), \bibinfo{booktitle}{Computer {Vision} – {ECCV} 2014}. \bibinfo{publisher}{Springer International Publishing}, \bibinfo{address}{Cham}. volume \bibinfo{volume}{8693}, pp. \bibinfo{pages}{740--755}.
\newblock \URLprefix \url{http://link.springer.com/10.1007/978-3-319-10602-1_48}, \DOIprefix\doi{10.1007/978-3-319-10602-1_48}. \bibinfo{note}{series Title: Lecture Notes in Computer Science}.
\bibitem[{Liu et~al.(2019)Liu, Ott, Goyal, Du, Joshi, Chen, Levy, Lewis, Zettlemoyer and Stoyanov}]{liu_roberta_2019}
\bibinfo{author}{Liu, Y.}, \bibinfo{author}{Ott, M.}, \bibinfo{author}{Goyal, N.}, \bibinfo{author}{Du, J.}, \bibinfo{author}{Joshi, M.}, \bibinfo{author}{Chen, D.}, \bibinfo{author}{Levy, O.}, \bibinfo{author}{Lewis, M.}, \bibinfo{author}{Zettlemoyer, L.}, \bibinfo{author}{Stoyanov, V.}, \bibinfo{year}{2019}.
\newblock \bibinfo{title}{{RoBERTa}: {A} {Robustly} {Optimized} {BERT} {Pretraining} {Approach}}.
\newblock \URLprefix \url{http://arxiv.org/abs/1907.11692}. \bibinfo{note}{arXiv:1907.11692 [cs]}.
\bibitem[{Lobry et~al.(2021)Lobry, Demir and Tuia}]{lobry_rsvqa_2021}
\bibinfo{author}{Lobry, S.}, \bibinfo{author}{Demir, B.}, \bibinfo{author}{Tuia, D.}, \bibinfo{year}{2021}.
\newblock \bibinfo{title}{{RSVQA} meets {BigEarthNet}: a new, large-scale, visual question answering dataset for remote sensing}, \bibinfo{publisher}{IEEE}.
\bibitem[{Lobry et~al.(2020a)Lobry, Marcos, Kellenberger and Tuia}]{lobry_better_2020}
\bibinfo{author}{Lobry, S.}, \bibinfo{author}{Marcos, D.}, \bibinfo{author}{Kellenberger, B.}, \bibinfo{author}{Tuia, D.}, \bibinfo{year}{2020}a.
\newblock \bibinfo{title}{Better {Generic} {Objects} {Counting} when {Asking} {Questions} to {Images}: {A} {Multitask} {Approach} for {Remote} {Sensing} {Visual} {Question} {Answering}}.
\newblock \bibinfo{journal}{ISPRS Annals of Photogrammetry, Remote Sensing and Spatial Information Sciences} \bibinfo{volume}{V-2-2020}, \bibinfo{pages}{1021--1027}.
\newblock \URLprefix \url{https://www.isprs-ann-photogramm-remote-sens-spatial-inf-sci.net/V-2-2020/1021/2020/}, \DOIprefix\doi{10.5194/isprs-annals-V-2-2020-1021-2020}.
\bibitem[{Lobry et~al.(2020b)Lobry, Marcos, Murray and Tuia}]{lobry_rsvqa_2020}
\bibinfo{author}{Lobry, S.}, \bibinfo{author}{Marcos, D.}, \bibinfo{author}{Murray, J.}, \bibinfo{author}{Tuia, D.}, \bibinfo{year}{2020}b.
\newblock \bibinfo{title}{{RSVQA}: {Visual} {Question} {Answering} for {Remote} {Sensing} {Data}}.
\newblock \bibinfo{journal}{IEEE Transactions on Geoscience and Remote Sensing} \bibinfo{volume}{58}, \bibinfo{pages}{8555--8566}.
\newblock \URLprefix \url{https://ieeexplore.ieee.org/document/9088993/}, \DOIprefix\doi{10.1109/TGRS.2020.2988782}.
\bibitem[{Mikolov et~al.(2013)Mikolov, Chen, Corrado and Dean}]{mikolov_efficient_2013}
\bibinfo{author}{Mikolov, T.}, \bibinfo{author}{Chen, K.}, \bibinfo{author}{Corrado, G.}, \bibinfo{author}{Dean, J.}, \bibinfo{year}{2013}.
\newblock \bibinfo{title}{Efficient {Estimation} of {Word} {Representations} in {Vector} {Space}}.
\newblock \URLprefix \url{http://arxiv.org/abs/1301.3781}. \bibinfo{note}{arXiv:1301.3781 [cs]}.
\bibitem[{Niu et~al.(2021)Niu, Tang, Zhang, Lu, Hua and Wen}]{niu_counterfactual_2021}
\bibinfo{author}{Niu, Y.}, \bibinfo{author}{Tang, K.}, \bibinfo{author}{Zhang, H.}, \bibinfo{author}{Lu, Z.}, \bibinfo{author}{Hua, X.S.}, \bibinfo{author}{Wen, J.R.}, \bibinfo{year}{2021}.
\newblock \bibinfo{title}{Counterfactual {VQA}: {A} {Cause}-{Effect} {Look} at {Language} {Bias}}, in: \bibinfo{booktitle}{2021 {IEEE}/{CVF} {Conference} on {Computer} {Vision} and {Pattern} {Recognition} ({CVPR})}, \bibinfo{publisher}{IEEE}, \bibinfo{address}{Nashville, TN, USA}. pp. \bibinfo{pages}{12695--12705}.
\newblock \URLprefix \url{https://ieeexplore.ieee.org/document/9578738/}, \DOIprefix\doi{10.1109/CVPR46437.2021.01251}.
\bibitem[{Ordonez et~al.(2011)Ordonez, Kulkarni and Berg}]{Ordonez_Im2Text_2011}
\bibinfo{author}{Ordonez, V.}, \bibinfo{author}{Kulkarni, G.}, \bibinfo{author}{Berg, T.}, \bibinfo{year}{2011}.
\newblock \bibinfo{title}{{Im2Text}: {Describing} {Images} {Using} 1 {Million} {Captioned} {Photographs}}, in: \bibinfo{editor}{Shawe-Taylor, J.}, \bibinfo{editor}{Zemel, R.}, \bibinfo{editor}{Bartlett, P.}, \bibinfo{editor}{Pereira, F.}, \bibinfo{editor}{Weinberger, K.Q.} (Eds.), \bibinfo{booktitle}{Advances in {Neural} {Information} {Processing} {Systems}}, \bibinfo{publisher}{Curran Associates, Inc.}
\newblock \URLprefix \url{https://proceedings.neurips.cc/paper/2011/file/5dd9db5e033da9c6fb5ba83c7a7ebea9-Paper.pdf}.
\bibitem[{Rahnemoonfar et~al.(2020)Rahnemoonfar, Chowdhury, Sarkar, Varshney, Yari and Murphy}]{rahnemoonfar_floodnet_2020}
\bibinfo{author}{Rahnemoonfar, M.}, \bibinfo{author}{Chowdhury, T.}, \bibinfo{author}{Sarkar, A.}, \bibinfo{author}{Varshney, D.}, \bibinfo{author}{Yari, M.}, \bibinfo{author}{Murphy, R.}, \bibinfo{year}{2020}.
\newblock \bibinfo{title}{{FloodNet}: {A} {High} {Resolution} {Aerial} {Imagery} {Dataset} for {Post} {Flood} {Scene} {Understanding}}.
\newblock \bibinfo{journal}{arXiv:2012.02951 [cs]} \URLprefix \url{http://arxiv.org/abs/2012.02951}. \bibinfo{note}{arXiv: 2012.02951}.
\bibitem[{Sanh et~al.(2020)Sanh, Debut, Chaumond and Wolf}]{sanh_distilbert_2020}
\bibinfo{author}{Sanh, V.}, \bibinfo{author}{Debut, L.}, \bibinfo{author}{Chaumond, J.}, \bibinfo{author}{Wolf, T.}, \bibinfo{year}{2020}.
\newblock \bibinfo{title}{{DistilBERT}, a distilled version of {BERT}: smaller, faster, cheaper and lighter}, in: \bibinfo{booktitle}{5th {Workshop} on {Energy} {Efficient} {Machine} {Learning} and {Cognitive} {Computing}}, \bibinfo{publisher}{arXiv}.
\newblock \URLprefix \url{http://arxiv.org/abs/1910.01108}. \bibinfo{note}{arXiv:1910.01108 [cs]}.
\bibitem[{Sharma et~al.(2018)Sharma, Ding, Goodman and Soricut}]{sharma_conceptual_2018}
\bibinfo{author}{Sharma, P.}, \bibinfo{author}{Ding, N.}, \bibinfo{author}{Goodman, S.}, \bibinfo{author}{Soricut, R.}, \bibinfo{year}{2018}.
\newblock \bibinfo{title}{Conceptual {Captions}: {A} {Cleaned}, {Hypernymed}, {Image} {Alt}-text {Dataset} {For} {Automatic} {Image} {Captioning}}, in: \bibinfo{booktitle}{Proceedings of the 56th {Annual} {Meeting} of the {Association} for {Computational} {Linguistics} ({Volume} 1: {Long} {Papers})}, \bibinfo{publisher}{Association for Computational Linguistics}, \bibinfo{address}{Melbourne, Australia}. pp. \bibinfo{pages}{2556--2565}.
\newblock \URLprefix \url{http://aclweb.org/anthology/P18-1238}, \DOIprefix\doi{10.18653/v1/P18-1238}.
\bibitem[{Siebert et~al.(2022)Siebert, Clasen, Ravanbakhsh and Demir}]{siebert_multi-modal_2022}
\bibinfo{author}{Siebert, T.}, \bibinfo{author}{Clasen, K.N.}, \bibinfo{author}{Ravanbakhsh, M.}, \bibinfo{author}{Demir, B.}, \bibinfo{year}{2022}.
\newblock \bibinfo{title}{Multi-modal fusion transformer for visual question answering in remote sensing}, in: \bibinfo{editor}{Pierdicca, N.}, \bibinfo{editor}{Bruzzone, L.}, \bibinfo{editor}{Bovolo, F.} (Eds.), \bibinfo{booktitle}{Image and {Signal} {Processing} for {Remote} {Sensing} {XXVIII}}, \bibinfo{publisher}{SPIE}, \bibinfo{address}{Berlin, Germany}. p.~\bibinfo{pages}{21}.
\newblock \URLprefix \url{https://www.spiedigitallibrary.org/conference-proceedings-of-spie/12267/2636276/Multi-modal-fusion-transformer-for-visual-question-answering-in-remote/10.1117/12.2636276.full}, \DOIprefix\doi{10.1117/12.2636276}.
\bibitem[{Sumbul et~al.(2019)Sumbul, Charfuelan, Demir and Markl}]{sumbul_bigearthnet_2019}
\bibinfo{author}{Sumbul, G.}, \bibinfo{author}{Charfuelan, M.}, \bibinfo{author}{Demir, B.}, \bibinfo{author}{Markl, V.}, \bibinfo{year}{2019}.
\newblock \bibinfo{title}{Bigearthnet: {A} {Large}-{Scale} {Benchmark} {Archive} for {Remote} {Sensing} {Image} {Understanding}}, in: \bibinfo{booktitle}{{IGARSS} 2019 - 2019 {IEEE} {International} {Geoscience} and {Remote} {Sensing} {Symposium}}, \bibinfo{publisher}{IEEE}, \bibinfo{address}{Yokohama, Japan}. pp. \bibinfo{pages}{5901--5904}.
\newblock \URLprefix \url{https://ieeexplore.ieee.org/document/8900532/}, \DOIprefix\doi{10.1109/IGARSS.2019.8900532}.
\bibitem[{Teney et~al.(2017)Teney, Liu and van~den Hengel}]{teney_graph-structured_2017}
\bibinfo{author}{Teney, D.}, \bibinfo{author}{Liu, L.}, \bibinfo{author}{van~den Hengel, A.}, \bibinfo{year}{2017}.
\newblock \bibinfo{title}{Graph-{Structured} {Representations} for {Visual} {Question} {Answering}}, pp. \bibinfo{pages}{1--9}.
\newblock \URLprefix \url{https://openaccess.thecvf.com/content_cvpr_2017/html/Teney_Graph-Structured_Representations_for_CVPR_2017_paper.html}.
\bibitem[{Tsimpoukelli et~al.(2021)Tsimpoukelli, Menick, Cabi, Eslami, Vinyals and Hill}]{tsimpoukelli_multimodal_2021}
\bibinfo{author}{Tsimpoukelli, M.}, \bibinfo{author}{Menick, J.}, \bibinfo{author}{Cabi, S.}, \bibinfo{author}{Eslami, S.M.A.}, \bibinfo{author}{Vinyals, O.}, \bibinfo{author}{Hill, F.}, \bibinfo{year}{2021}.
\newblock \bibinfo{title}{Multimodal {Few}-{Shot} {Learning} with {Frozen} {Language} {Models}}.
\newblock \bibinfo{journal}{Advances in Neural Information Processing Systems} \bibinfo{volume}{34}.
\newblock \URLprefix \url{https://papers.nips.cc/paper/2021/hash/01b7575c38dac42f3cfb7d500438b875-Abstract.html}.
\bibitem[{Tuia et~al.(2021)Tuia, Roscher, Wegner, Jacobs, Zhu and Camps-Valls}]{tuia_towards_2021}
\bibinfo{author}{Tuia, D.}, \bibinfo{author}{Roscher, R.}, \bibinfo{author}{Wegner, J.D.}, \bibinfo{author}{Jacobs, N.}, \bibinfo{author}{Zhu, X.X.}, \bibinfo{author}{Camps-Valls, G.}, \bibinfo{year}{2021}.
\newblock \bibinfo{title}{Towards a {Collective} {Agenda} on {AI} for {Earth} {Science} {Data} {Analysis}}.
\newblock \bibinfo{journal}{IEEE Geoscience and Remote Sensing Magazine} \bibinfo{volume}{9}, \bibinfo{pages}{88--104}.
\newblock \URLprefix \url{http://arxiv.org/abs/2104.05107}, \DOIprefix\doi{10.1109/MGRS.2020.3043504}. \bibinfo{note}{arXiv: 2104.05107}.
\bibitem[{Vaswani et~al.(2017)Vaswani, Shazeer, Parmar, Uszkoreit, Jones, Gomez, Kaiser and Polosukhin}]{vaswani_attention_2017}
\bibinfo{author}{Vaswani, A.}, \bibinfo{author}{Shazeer, N.}, \bibinfo{author}{Parmar, N.}, \bibinfo{author}{Uszkoreit, J.}, \bibinfo{author}{Jones, L.}, \bibinfo{author}{Gomez, A.N.}, \bibinfo{author}{Kaiser, L.}, \bibinfo{author}{Polosukhin, I.}, \bibinfo{year}{2017}.
\newblock \bibinfo{title}{Attention is {All} you {Need}}, in: \bibinfo{editor}{Guyon, I.}, \bibinfo{editor}{Luxburg, U.V.}, \bibinfo{editor}{Bengio, S.}, \bibinfo{editor}{Wallach, H.}, \bibinfo{editor}{Fergus, R.}, \bibinfo{editor}{Vishwanathan, S.}, \bibinfo{editor}{Garnett, R.} (Eds.), \bibinfo{booktitle}{Advances in {Neural} {Information} {Processing} {Systems}}, \bibinfo{publisher}{Curran Associates, Inc.}
\newblock \URLprefix \url{https://proceedings.neurips.cc/paper/2017/file/3f5ee243547dee91fbd053c1c4a845aa-Paper.pdf}.
\bibitem[{Wang et~al.(2021)Wang, Hu, Gan, Yang, Dai, Liu, Lu and Wang}]{wang_ufo_2021}
\bibinfo{author}{Wang, J.}, \bibinfo{author}{Hu, X.}, \bibinfo{author}{Gan, Z.}, \bibinfo{author}{Yang, Z.}, \bibinfo{author}{Dai, X.}, \bibinfo{author}{Liu, Z.}, \bibinfo{author}{Lu, Y.}, \bibinfo{author}{Wang, L.}, \bibinfo{year}{2021}.
\newblock \bibinfo{title}{{UFO}: {A} {UniFied} {TransfOrmer} for {Vision}-{Language} {Representation} {Learning}}.
\newblock \bibinfo{journal}{arXiv:2111.10023 [cs]} \URLprefix \url{http://arxiv.org/abs/2111.10023}. \bibinfo{note}{arXiv: 2111.10023}.
\bibitem[{Wolf et~al.(2020)Wolf, Debut, Sanh, Chaumond, Delangue, Moi, Cistac, Rault, Louf, Funtowicz, Davison, Shleifer, von Platen, Ma, Jernite, Plu, Xu, Scao, Gugger, Drame, Lhoest and Rush}]{wolf_huggingfaces_2020}
\bibinfo{author}{Wolf, T.}, \bibinfo{author}{Debut, L.}, \bibinfo{author}{Sanh, V.}, \bibinfo{author}{Chaumond, J.}, \bibinfo{author}{Delangue, C.}, \bibinfo{author}{Moi, A.}, \bibinfo{author}{Cistac, P.}, \bibinfo{author}{Rault, T.}, \bibinfo{author}{Louf, R.}, \bibinfo{author}{Funtowicz, M.}, \bibinfo{author}{Davison, J.}, \bibinfo{author}{Shleifer, S.}, \bibinfo{author}{von Platen, P.}, \bibinfo{author}{Ma, C.}, \bibinfo{author}{Jernite, Y.}, \bibinfo{author}{Plu, J.}, \bibinfo{author}{Xu, C.}, \bibinfo{author}{Scao, T.L.}, \bibinfo{author}{Gugger, S.}, \bibinfo{author}{Drame, M.}, \bibinfo{author}{Lhoest, Q.}, \bibinfo{author}{Rush, A.M.}, \bibinfo{year}{2020}.
\newblock \bibinfo{title}{{HuggingFace}'s {Transformers}: {State}-of-the-art {Natural} {Language} {Processing}}.
\newblock \URLprefix \url{http://arxiv.org/abs/1910.03771}. \bibinfo{note}{arXiv:1910.03771 [cs]}.
\bibitem[{Wu et~al.(2018)Wu, Shen, Wang, Dick and Hengel}]{wu_image_2018}
\bibinfo{author}{Wu, Q.}, \bibinfo{author}{Shen, C.}, \bibinfo{author}{Wang, P.}, \bibinfo{author}{Dick, A.}, \bibinfo{author}{Hengel, A.v.d.}, \bibinfo{year}{2018}.
\newblock \bibinfo{title}{Image {Captioning} and {Visual} {Question} {Answering} {Based} on {Attributes} and {External} {Knowledge}}.
\newblock \bibinfo{journal}{IEEE Transactions on Pattern Analysis and Machine Intelligence} \bibinfo{volume}{40}, \bibinfo{pages}{1367--1381}.
\newblock \DOIprefix\doi{10.1109/TPAMI.2017.2708709}. \bibinfo{note}{conference Name: IEEE Transactions on Pattern Analysis and Machine Intelligence}.
\bibitem[{Wu et~al.(2016)Wu, Schuster, Chen, Le, Norouzi, Macherey, Krikun, Cao, Gao, Macherey, Klingner, Shah, Johnson, Liu, Kaiser, Gouws, Kato, Kudo, Kazawa, Stevens, Kurian, Patil, Wang, Young, Smith, Riesa, Rudnick, Vinyals, Corrado, Hughes and Dean}]{wu_googles_2016}
\bibinfo{author}{Wu, Y.}, \bibinfo{author}{Schuster, M.}, \bibinfo{author}{Chen, Z.}, \bibinfo{author}{Le, Q.V.}, \bibinfo{author}{Norouzi, M.}, \bibinfo{author}{Macherey, W.}, \bibinfo{author}{Krikun, M.}, \bibinfo{author}{Cao, Y.}, \bibinfo{author}{Gao, Q.}, \bibinfo{author}{Macherey, K.}, \bibinfo{author}{Klingner, J.}, \bibinfo{author}{Shah, A.}, \bibinfo{author}{Johnson, M.}, \bibinfo{author}{Liu, X.}, \bibinfo{author}{Kaiser, L.}, \bibinfo{author}{Gouws, S.}, \bibinfo{author}{Kato, Y.}, \bibinfo{author}{Kudo, T.}, \bibinfo{author}{Kazawa, H.}, \bibinfo{author}{Stevens, K.}, \bibinfo{author}{Kurian, G.}, \bibinfo{author}{Patil, N.}, \bibinfo{author}{Wang, W.}, \bibinfo{author}{Young, C.}, \bibinfo{author}{Smith, J.}, \bibinfo{author}{Riesa, J.}, \bibinfo{author}{Rudnick, A.}, \bibinfo{author}{Vinyals, O.}, \bibinfo{author}{Corrado, G.}, \bibinfo{author}{Hughes, M.}, \bibinfo{author}{Dean, J.}, \bibinfo{year}{2016}.
\newblock \bibinfo{title}{Google's {Neural} {Machine} {Translation} {System}: {Bridging} the {Gap} between {Human} and {Machine} {Translation}}.
\newblock \bibinfo{journal}{arXiv:1609.08144 [cs]} \URLprefix \url{http://arxiv.org/abs/1609.08144}. \bibinfo{note}{arXiv: 1609.08144}.
\bibitem[{Xia et~al.(2018)Xia, Bai, Ding, Zhu, Belongie, Luo, Datcu, Pelillo and Zhang}]{xia_dota_2018}
\bibinfo{author}{Xia, G.S.}, \bibinfo{author}{Bai, X.}, \bibinfo{author}{Ding, J.}, \bibinfo{author}{Zhu, Z.}, \bibinfo{author}{Belongie, S.}, \bibinfo{author}{Luo, J.}, \bibinfo{author}{Datcu, M.}, \bibinfo{author}{Pelillo, M.}, \bibinfo{author}{Zhang, L.}, \bibinfo{year}{2018}.
\newblock \bibinfo{title}{{DOTA}: {A} {Large}-{Scale} {Dataset} for {Object} {Detection} in {Aerial} {Images}}, in: \bibinfo{booktitle}{2018 {IEEE}/{CVF} {Conference} on {Computer} {Vision} and {Pattern} {Recognition}}, \bibinfo{publisher}{IEEE}, \bibinfo{address}{Salt Lake City, UT}. pp. \bibinfo{pages}{3974--3983}.
\newblock \URLprefix \url{https://ieeexplore.ieee.org/document/8578516/}, \DOIprefix\doi{10.1109/CVPR.2018.00418}.
\bibitem[{Xia et~al.(2017)Xia, Hu, Hu, Shi, Bai, Zhong, Zhang and Lu}]{xia_aid_2017}
\bibinfo{author}{Xia, G.S.}, \bibinfo{author}{Hu, J.}, \bibinfo{author}{Hu, F.}, \bibinfo{author}{Shi, B.}, \bibinfo{author}{Bai, X.}, \bibinfo{author}{Zhong, Y.}, \bibinfo{author}{Zhang, L.}, \bibinfo{author}{Lu, X.}, \bibinfo{year}{2017}.
\newblock \bibinfo{title}{{AID}: {A} {Benchmark} {Data} {Set} for {Performance} {Evaluation} of {Aerial} {Scene} {Classification}}.
\newblock \bibinfo{journal}{IEEE Transactions on Geoscience and Remote Sensing} \bibinfo{volume}{55}, \bibinfo{pages}{3965--3981}.
\newblock \URLprefix \url{http://ieeexplore.ieee.org/document/7907303/}, \DOIprefix\doi{10.1109/TGRS.2017.2685945}.
\bibitem[{Yang and Newsam(2010)}]{yang_bag--visual-words_2010}
\bibinfo{author}{Yang, Y.}, \bibinfo{author}{Newsam, S.}, \bibinfo{year}{2010}.
\newblock \bibinfo{title}{Bag-of-visual-words and spatial extensions for land-use classification}, in: \bibinfo{booktitle}{Proceedings of the 18th {SIGSPATIAL} {International} {Conference} on {Advances} in {Geographic} {Information} {Systems} - {GIS} '10}, \bibinfo{publisher}{ACM Press}, \bibinfo{address}{San Jose, California}. p. \bibinfo{pages}{270}.
\newblock \URLprefix \url{http://portal.acm.org/citation.cfm?doid=1869790.1869829}, \DOIprefix\doi{10.1145/1869790.1869829}.
\bibitem[{Yang et~al.(2016)Yang, He, Gao, Deng and Smola}]{yang_stacked_2016}
\bibinfo{author}{Yang, Z.}, \bibinfo{author}{He, X.}, \bibinfo{author}{Gao, J.}, \bibinfo{author}{Deng, L.}, \bibinfo{author}{Smola, A.}, \bibinfo{year}{2016}.
\newblock \bibinfo{title}{Stacked {Attention} {Networks} for {Image} {Question} {Answering}}, in: \bibinfo{booktitle}{2016 {IEEE} {Conference} on {Computer} {Vision} and {Pattern} {Recognition} ({CVPR})}, \bibinfo{publisher}{IEEE}, \bibinfo{address}{Las Vegas, NV, USA}. pp. \bibinfo{pages}{21--29}.
\newblock \URLprefix \url{http://ieeexplore.ieee.org/document/7780379/}, \DOIprefix\doi{10.1109/CVPR.2016.10}.
\bibitem[{Yu et~al.(2022)Yu, Wang, Vasudevan, Yeung, Seyedhosseini and Wu}]{yu_coca_2022}
\bibinfo{author}{Yu, J.}, \bibinfo{author}{Wang, Z.}, \bibinfo{author}{Vasudevan, V.}, \bibinfo{author}{Yeung, L.}, \bibinfo{author}{Seyedhosseini, M.}, \bibinfo{author}{Wu, Y.}, \bibinfo{year}{2022}.
\newblock \bibinfo{title}{{CoCa}: {Contrastive} {Captioners} are {Image}-{Text} {Foundation} {Models}}.
\newblock \bibinfo{journal}{Transactions on Machine Learning Research} \URLprefix \url{https://openreview.net/forum?id=Ee277P3AYC}.
\bibitem[{Yuan et~al.(2021)Yuan, Chen, Chen, Codella, Dai, Gao, Hu, Huang, Li, Li, Liu, Liu, Liu, Lu, Shi, Wang, Wang, Xiao, Xiao, Yang, Zeng, Zhou and Zhang}]{yuan_florence_2021}
\bibinfo{author}{Yuan, L.}, \bibinfo{author}{Chen, D.}, \bibinfo{author}{Chen, Y.L.}, \bibinfo{author}{Codella, N.}, \bibinfo{author}{Dai, X.}, \bibinfo{author}{Gao, J.}, \bibinfo{author}{Hu, H.}, \bibinfo{author}{Huang, X.}, \bibinfo{author}{Li, B.}, \bibinfo{author}{Li, C.}, \bibinfo{author}{Liu, C.}, \bibinfo{author}{Liu, M.}, \bibinfo{author}{Liu, Z.}, \bibinfo{author}{Lu, Y.}, \bibinfo{author}{Shi, Y.}, \bibinfo{author}{Wang, L.}, \bibinfo{author}{Wang, J.}, \bibinfo{author}{Xiao, B.}, \bibinfo{author}{Xiao, Z.}, \bibinfo{author}{Yang, J.}, \bibinfo{author}{Zeng, M.}, \bibinfo{author}{Zhou, L.}, \bibinfo{author}{Zhang, P.}, \bibinfo{year}{2021}.
\newblock \bibinfo{title}{Florence: {A} {New} {Foundation} {Model} for {Computer} {Vision}}.
\newblock \bibinfo{journal}{arXiv:2111.11432 [cs]} \URLprefix \url{http://arxiv.org/abs/2111.11432}. \bibinfo{note}{arXiv: 2111.11432}.
\bibitem[{Yuan et~al.(2022a)Yuan, Mou, Wang and Zhu}]{yuan_easy_2022}
\bibinfo{author}{Yuan, Z.}, \bibinfo{author}{Mou, L.}, \bibinfo{author}{Wang, Q.}, \bibinfo{author}{Zhu, X.X.}, \bibinfo{year}{2022}a.
\newblock \bibinfo{title}{From {Easy} to {Hard}: {Learning} {Language}-guided {Curriculum} for {Visual} {Question} {Answering} on {Remote} {Sensing} {Data}}.
\newblock \bibinfo{journal}{IEEE Transactions on Geoscience and Remote Sensing} \bibinfo{volume}{60}, \bibinfo{pages}{1--11}.
\newblock \URLprefix \url{http://arxiv.org/abs/2205.03147}, \DOIprefix\doi{10.1109/TGRS.2022.3173811}. \bibinfo{note}{arXiv:2205.03147 [cs]}.
\bibitem[{Yuan et~al.(2022b)Yuan, Mou, Xiong and Zhu}]{yuan_change_2022}
\bibinfo{author}{Yuan, Z.}, \bibinfo{author}{Mou, L.}, \bibinfo{author}{Xiong, Z.}, \bibinfo{author}{Zhu, X.X.}, \bibinfo{year}{2022}b.
\newblock \bibinfo{title}{Change {Detection} {Meets} {Visual} {Question} {Answering}}.
\newblock \bibinfo{journal}{IEEE Transactions on Geoscience and Remote Sensing} \bibinfo{volume}{60}, \bibinfo{pages}{1--13}.
\newblock \URLprefix \url{https://ieeexplore.ieee.org/document/9901476/}, \DOIprefix\doi{10.1109/TGRS.2022.3203314}.
\bibitem[{Zellers et~al.(2021)Zellers, Lu, Hessel, Yu, Park, Cao, Farhadi and Choi}]{zellers_merlot_2021}
\bibinfo{author}{Zellers, R.}, \bibinfo{author}{Lu, X.}, \bibinfo{author}{Hessel, J.}, \bibinfo{author}{Yu, Y.}, \bibinfo{author}{Park, J.S.}, \bibinfo{author}{Cao, J.}, \bibinfo{author}{Farhadi, A.}, \bibinfo{author}{Choi, Y.}, \bibinfo{year}{2021}.
\newblock \bibinfo{title}{{MERLOT}: {Multimodal} {Neural} {Script} {Knowledge} {Models}}.
\newblock \bibinfo{journal}{arXiv:2106.02636 [cs]} \URLprefix \url{http://arxiv.org/abs/2106.02636}. \bibinfo{note}{arXiv: 2106.02636}.
\bibitem[{Zhang et~al.(2015)Zhang, Du and Zhang}]{zhang_saliency-guided_2015}
\bibinfo{author}{Zhang, F.}, \bibinfo{author}{Du, B.}, \bibinfo{author}{Zhang, L.}, \bibinfo{year}{2015}.
\newblock \bibinfo{title}{Saliency-{Guided} {Unsupervised} {Feature} {Learning} for {Scene} {Classification}}.
\newblock \bibinfo{journal}{IEEE Transactions on Geoscience and Remote Sensing} \bibinfo{volume}{53}, \bibinfo{pages}{2175--2184}.
\newblock \URLprefix \url{http://ieeexplore.ieee.org/document/6910306/}, \DOIprefix\doi{10.1109/TGRS.2014.2357078}.
\bibitem[{Zhang et~al.(2019)Zhang, Yuan, Feng and Lu}]{zhang_hierarchical_2019}
\bibinfo{author}{Zhang, Y.}, \bibinfo{author}{Yuan, Y.}, \bibinfo{author}{Feng, Y.}, \bibinfo{author}{Lu, X.}, \bibinfo{year}{2019}.
\newblock \bibinfo{title}{Hierarchical and {Robust} {Convolutional} {Neural} {Network} for {Very} {High}-{Resolution} {Remote} {Sensing} {Object} {Detection}}.
\newblock \bibinfo{journal}{IEEE Transactions on Geoscience and Remote Sensing} \bibinfo{volume}{57}, \bibinfo{pages}{5535--5548}.
\newblock \URLprefix \url{https://ieeexplore.ieee.org/document/8676107/}, \DOIprefix\doi{10.1109/TGRS.2019.2900302}.
\bibitem[{Zheng et~al.(2021)Zheng, Wang, Du and Lu}]{zheng_mutual_2021}
\bibinfo{author}{Zheng, X.}, \bibinfo{author}{Wang, B.}, \bibinfo{author}{Du, X.}, \bibinfo{author}{Lu, X.}, \bibinfo{year}{2021}.
\newblock \bibinfo{title}{Mutual {Attention} {Inception} {Network} for {Remote} {Sensing} {Visual} {Question} {Answering}}.
\newblock \bibinfo{journal}{IEEE Transactions on Geoscience and Remote Sensing} , \bibinfo{pages}{1--14}\URLprefix \url{https://ieeexplore.ieee.org/document/9444570/}, \DOIprefix\doi{10.1109/TGRS.2021.3079918}.

\end{thebibliography}

\newpage
\onecolumn
\appendix
\section{Appendix}

\begin{figure}[!h]
\centering
\includegraphics[width=0.5\textwidth]{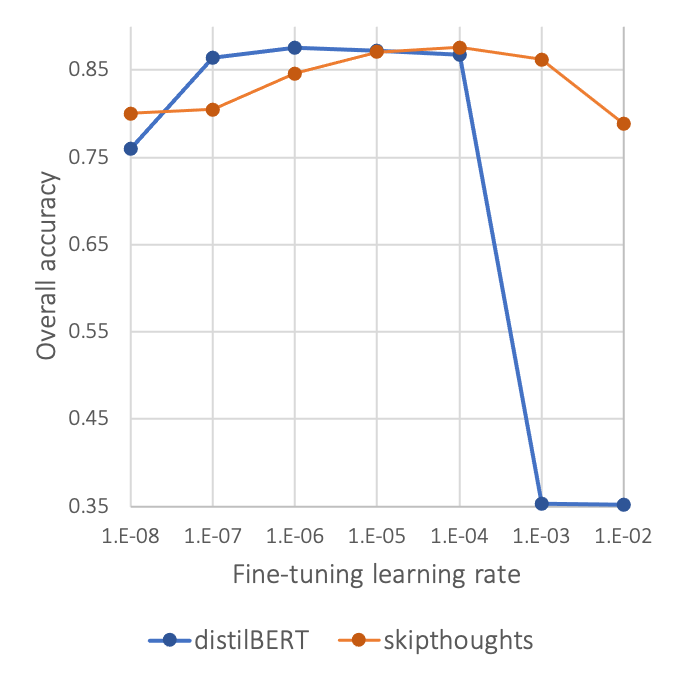}
\caption{Hyper-parameter search on the validation set of RSVQA Low Resolution for the fine-tuning learning rates of the language models, Skip-thoughts RNN and distilBERT Transformer.}
\label{fig:FineTuneLearningRate}
\end{figure}

\begin{table}[H]
\caption{Numerical results of the visual blind models and adversarial testing, compared to the test results for the reference models, that are provided with both question and image (averaged over three runs, with standard deviation in brackets). All of these models have an element-wise fusion to remove as much additional complexity and give the focus to the assessment of language.
A difference is systematically calculated between the disturbed setup (visual blind or adversarial testing) and the corresponding reference model averaged results. 
}
\label{tab:blindAdResultsLR}
\centering
\resizebox{0.99\linewidth}{!}{%
\begin{tabular}{l|l|c|c|c|c|c|c|c|}
\hhline{~~~------}
\multicolumn{3}{c|}{} & \multicolumn{6}{c|}{Accuracy}\\
Language model & & & Comparison & Counting & Presence & Rural/urban & AA & OA \\
\hline
\multirow{10}{*}{RNN Skip-thoughts} & \multirow{5}{*}{frozen} & Reference model &  79.89 (0.31) & 65.80 (0.32) & 87.57 (0.06) & 90.67 (0.94) & 80.98 (0.08) & 78.12 (0.21)\\
\hhline{~~-------}
&& Visual blind & 79.81 (0.43) & 57.30 (0.17) & 86.38 (0.03) & 56.00 (0.00) & 69.88 (0.15) & 74.89 (0.22)   \\
&& Difference & 0.08 & 8.5 & 1.19 & 34.67 & 11.1 & 3.23 \\
\hhline{~~-------}
&& Adversarial & 78.32 (0.27) & 41.42 (1.15) & 79.02 (0.27) & 56.00 (0.00) & 63.69 (0.28) & 67.43 (0.35) \\
&& Difference & 1.57 & 24.38 & 8.55 & 34.67 & 17.29 & 10.69 \\
\hhline{~--------}
\hhline{~--------}
\hhline{~--------}
& \multirow{5}{*}{fine-tuned} & Reference model & 90.59 (0.08) & 70.82 (0.46) & 90.51 (0.06) & 89.33 (4.50) & 85.31 (1.12) & 84.73 (0.16)\\
\hhline{~~-------}
&& Visual blind & 89.19 (0.15) & 60.91 (0.06) & 89.19 (0.25) & 56.00 (0.00) & 73.82 (0.06) & 80.53 (0.08)\\
&& Difference & 1.4 & 9.91 & 1.32 & 33.33 & 11.49 & 4.2 \\
\hhline{~~-------}
&& Adversarial & 78.55 (3.84) & 36.75 (3.43) & 70.89 (3.85) & 56.00 (0.00) & 60.55 (2.76) & 63.75 (3.66) \\
&& Difference & 12.04 & 34.07 & 19.62 & 33.33 & 24.76 & 20.98 \\
\hline
\hline
\multirow{10}{*}{Transformer distilBERT} & \multirow{5}{*}{frozen} & Reference model &  67.82 (0.10) & 63.16 (0.32) & 87.31 (0.24) & 87.67 (0.47) & 76.49 (0.29) & 73.42 (0.15)\\
\hhline{~~-------}
&& Visual blind & 68.39 (0.61) & 54.80 (0.29) & 86.63 (0.15) & 56.00 (0.00) & 66.46 (0.16) & 69.65 (0.24) \\
&& Difference & -0.57 & 8.36 & 0.68 & 31.67 & 10.03 & 3.77\\
\hhline{~~-------}
&& Adversarial & 67.76 (0.52) & 38.32 (1.26) & 76.53 (5.14) & 56.00 (0.00) & 59.65 (1.70) & 61.56 (2.06) \\
&& Difference & 0.06 & 24.84 & 10.78 & 31.67 & 16.84 & 11.86 \\
\hhline{~--------}
\hhline{~--------}
\hhline{~--------}
& \multirow{5}{*}{fine-tuned} & Reference model & 90.76 (0.26) & 70.59 (0.36) & 91.04 (0.30) & 90.00 (1.41) & 85.60 (0.36) & 84.90 (0.05)\\
\hhline{~~-------}
&& Visual blind & 89.60 (0.12) & 60.48 (0.54) & 89.58 (0.06) & 56.00 (0.00) & 73.91 (0.16) & 80.67 (0.20) \\
&& Difference & 1.16 & 10.11 & 1.46 & 34 & 11.69 & 4.23 \\
\hhline{~~-------}
&& Adversarial & 79.54 (3.94) & 38.87 (3.86) & 73.02 (3.21) & 56.00 (0.00) & 61.86 (2.72) & 65.40 (3.62) \\
&& Difference & 11.22 & 31.72 & 18.02 & 34 & 23.74 & 19.50 \\
\hline
\end{tabular}}
\end{table}

\begin{table}[!h]
    \caption{FloodNet - Low-level analysis - accuracy by predicting the most common answer in for each question.}
    \label{tab:FloodNet_low_level_analysis}
    \centering
    \resizebox{0.9\linewidth}{!}{%
    \begin{tabular}{|c|c|c|c|}
     \hline
     Question & Most common answer & Number of samples & $\mathbf{Prior\_Acc}$ (\%)\\
     \hline
     How many non flooded buildings can be seen in this image? & 3 & 179 & 24.58\\
     How many buildings can be seen in the image? & 3 & 151 & 20.53\\
     How many buildings can be seen in this image? & 4 & 173 & 18.49\\
     How many buildings are non flooded? & 4 & 183 & 23.50\\
     How many buildings are in this image? & 1 & 169 & 17.75\\
     How many buildings are non flooded in this image? & 1 & 179 & 25.14\\
     How many buildings are in the image? & 3 & 143 & 20.98\\
     How many buildings are flooded in this image? & 8 & 48 & 10.42\\
     How many buildings are flooded? & 17 & 49 & 10.20\\
     How many flooded buildings can be seen in this image? & 16 & 55 & 12.73\\
     Is the entire road flooded? & no & 426 & 82.16\\
     Is the entire road non flooded? & yes & 441 & 81.41\\
     What is the condition of road? & non flooded & 452 & 82.08\\
     What is the condition of the road in this image? & non flooded & 415 & 80.00\\
     What is the overall condition of the given image? & non flooded & 1448 & 86.12\\
     \hline
     Overall & & & \textbf{64.98} \\
     \hline
    \end{tabular}}
\end{table}

\end{document}